\documentclass[10pt,twocolumn,letterpaper]{article}

\usepackage{cvpr}              

\newcommand{\ts}{\textsuperscript}
\newcommand*{\td}{\textsuperscript{\textdagger}}

\usepackage{colortbl}
\usepackage{makecell}
\usepackage{multirow}
\usepackage[accsupp]{axessibility}  

\definecolor{1st}{RGB}{255,100,100}
\definecolor{2nd}{RGB}{248,143,59}
\definecolor{3rd}{RGB}{0,0,0}

\DeclareMathSizes{10}{8}{7}{5}

\definecolor{cvprblue}{rgb}{0.21,0.49,0.74}
\usepackage[pagebackref,breaklinks,colorlinks,allcolors=cvprblue]{hyperref}

\def\paperID{xxxx} 
\def\confName{CVPR}
\def\confYear{2026}

\title{2D Triangle Splatting for Direct Differentiable Mesh Training}

\author{
Kaifeng Sheng\footnotemark[1] \footnotemark[2] , Zheng Zhou\footnotemark[1] , Yingliang Peng, Qianwei Wang \\
Amap, Alibaba Group\\
{\tt\small kaifeng.skf@gmail.com, \{zhouzheng.zhou, yingliang.pyl, qianwei.wang\}@alibaba-inc.com}\\
}

\begin{document}
\maketitle

\renewcommand*{\thefootnote}{\fnsymbol{footnote}}
\footnotetext[1]{Equal contribution.}
\footnotetext[2]{Corresponding author.}
\renewcommand*{\thefootnote}{\arabic{footnote}}

\begin{abstract}
    Differentiable rendering with 3D Gaussian primitives has emerged as a powerful method for reconstructing high-fidelity 3D scenes from multi-view images.
    While it offers improvements over NeRF-based methods, this representation still encounters challenges with rendering speed and advanced rendering effects, such as relighting and shadow rendering, compared to mesh-based models.
    In this paper, we propose 2D Triangle Splatting (2DTS), a novel method that replaces 3D Gaussian primitives with 2D triangle primitives.
    This representation naturally forms a discrete mesh-like structure while retaining the benefits of continuous volumetric modeling.
    Through the incorporation and controlled annealing of a compactness parameter, our method maintains differentiability during training while producing triangle meshes with fully opaque faces at the end of optimization without the need for additional post-processing.
    Experimental results demonstrate that our triangle-based representation achieves competitive visual quality with Gaussian-based methods while providing a more direct bridge to mesh-based representations.
    Our method bridges the gap between differentiable rendering and traditional mesh-based rendering, offering a promising solution for applications requiring renderable mesh-like reconstructions.
\end{abstract}
\section{Introduction}
\label{sec:intro}

\begin{figure}[ht]
  \centering
  \includegraphics[width=0.99\columnwidth]{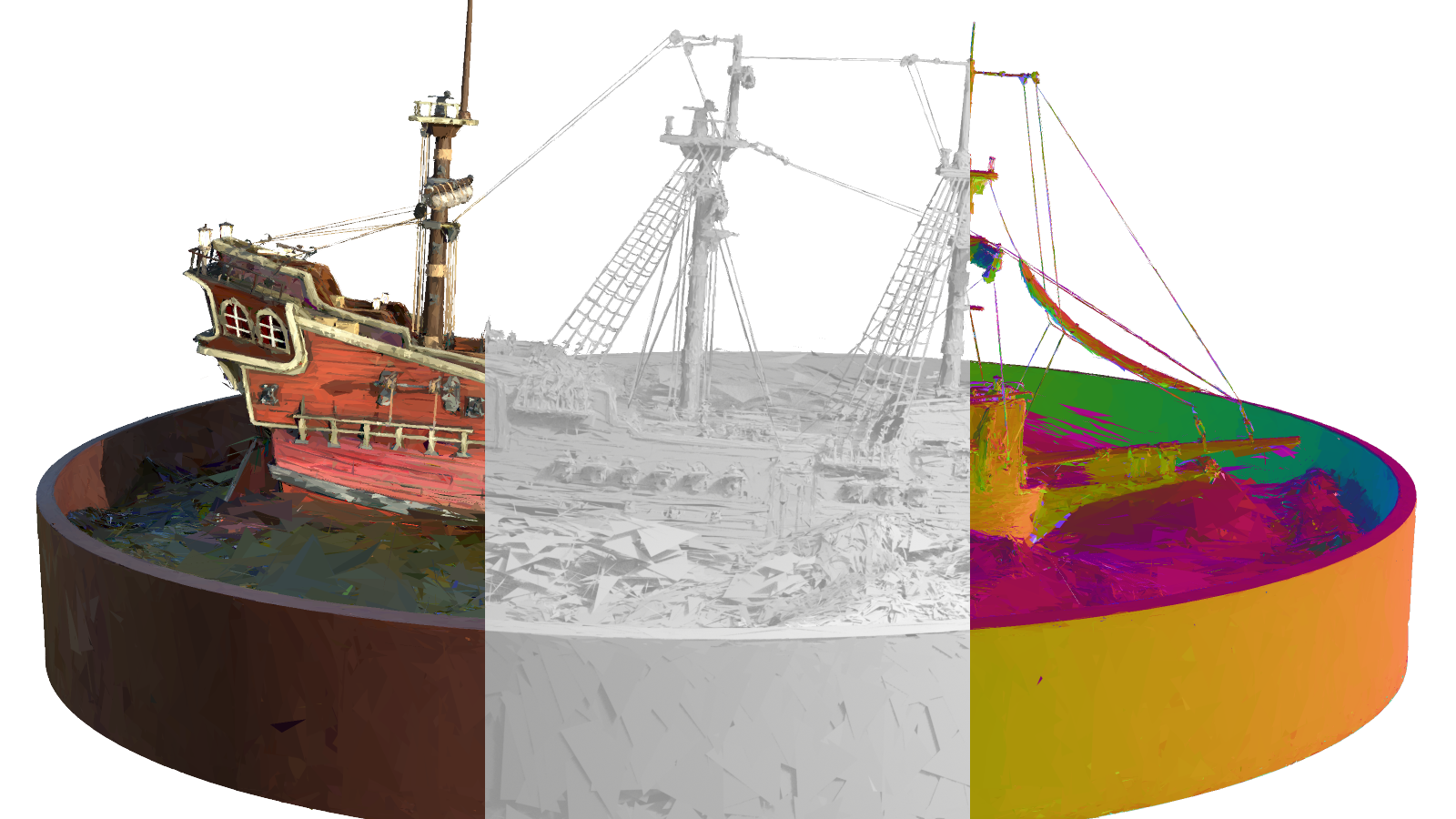}
  \caption{
    {\bf 2D Triangle Splatting for Mesh Reconstruction.}
      Visualization of a model optimized using our proposed 2D Triangle Splatting method on the NeRF-Synthetic \textit{Ship} scene.
  }
  \label{fig:demo}
\end{figure}

Reconstructing consistent 3D geometry and synthesizing novel views from multi-view images has been a long-standing challenge in computer vision. 
Traditional Multi-View Stereo (MVS) pipelines~\cite{Schonberger2016} generate point clouds by identifying correspondences between images and then reconstructing textured meshes from these point clouds.
However, the accuracy of these methods is limited by the quality of detected correspondences and suffers from issues with non-Lambertian surfaces and textureless regions.
They also tend to produce over-smoothed surfaces and struggle to represent fine structures due to limitations in the triangulation process.

Differentiable rendering and back-propagation-based optimization have advanced the field significantly.
Neural radiance fields (NeRF)~\cite{mildenhall2020nerf} represent geometry with a continuous function approximated by neural networks, improving upon traditional mesh-based methods in the visual quality of synthesized novel views. 
However, their applications are limited by high computational costs and the lack of explicit geometry representation. 
3D Gaussian splatting (3DGS)~\cite{kerbl3Dgaussians} addresses these limitations by representing geometry with 3D Gaussian primitives and implementing an efficient splatting method for differentiable rendering.
3DGS has proven more efficient than NeRF-based methods and can potentially achieve real-time rendering~\cite{lin2024metasapiens}.
However, the geometry from 3DGS still deviates from the actual scene geometry due to the diffuse nature of Gaussian kernels.
Recent works have improved 3DGS representation by rendering more accurate depth~\cite{Chen2024PGSR, zhang2024RaDe-GS} or replacing 3D Gaussians with primitives that have better geometric properties, such as 2D Gaussian plates~\cite{Huang2DGS2024} or 3D convexes~\cite{Held20243DConvex}.

Despite these improvements, splatting-based models are far less widely used than mesh models in real-world applications.
One key reason is that splatting-based models require specially designed rendering pipelines and remain incompatible with existing rendering engines. 
Although there have been recent attempts~\cite{Gao2024Relightable}, achieving advanced rendering effects, such as relighting and shadow rendering, with splatting-based models remains challenging.
The computationally intensive depth sorting and alpha-blending operations used in splatting-based methods also hinder their deployment on resource-constrained devices.
To extract triangle meshes from splatting-based models, a post-processing step using Truncated Signed Distance Function (TSDF)~\cite{Zhou2018, Richard2011KinectFusion} is typically required. 
This post-processing step is not end-to-end and may introduce artifacts in the final mesh. 
It also reduces the Lagrangian representation of Gaussian primitives to a voxel-based Eulerian representation, creating a trade-off between computational complexity and geometric fidelity~\cite{guo2025tetsphere}.
Compared to their original representation, meshes extracted from Gaussian-based methods are either too large or too coarse to efficiently represent different levels of detail in the scene.
A method that combines the portability and efficiency of meshes with the visual fidelity of radiance fields is therefore desirable.

In this work, we propose a novel approach that replaces the Gaussian primitives in 3DGS with 2D triangle primitives. 
These triangle primitives feature spatially varying opacity based on the barycentric coordinates of each point on the triangle plane, preserving differentiability with respect to vertex locations throughout the rendering process.
We demonstrate that by replacing 3D Gaussians with 2D triangles, we achieve strong visual quality compared to representative Gaussian-based methods in our experiments~\cite{kerbl3Dgaussians, Yu2023MipSplatting}.
Figure \ref{fig:demo} shows the visualization of a model optimized using our proposed method on the NeRF-Synthetic \textit{Ship} scene.
It can be observed that our method represents the geometry with discrete triangle faces while preserving the fine structures of the original scene, which are hard to reconstruct using traditional mesh reconstruction methods.

To bridge the gap between diffuse training-time triangles and a final mesh with opaque faces, we introduce a compactness parameter inspired by Generalized Exponential Splatting (GES)~\cite{hamdi_2024_CVPR}.
This parameter controls the sharpness of primitive edges by replacing the exponential decay function of 3DGS with a Generalized Exponential Function (GEF).
Rather than learning the compactness parameter, we employ a training schedule that gradually increases triangle compactness during training, yielding a mesh representation with fully opaque triangle faces at the end of training.
Because our entire training process is end-to-end without requiring post-processing, our method produces meshes with strong rendered visual quality, as demonstrated in our experiments.
Although the meshes generated by our method are not watertight and may contain overlaps between disconnected triangles, they preserve thin structures and rendered visual quality better than meshes extracted using TSDF-based methods in our experiments.
Our approach thus offers an attractive solution for rendering-focused applications such as 3D map reconstruction and virtual reality.

\section{Related Work}
\label{sec:related}

The field of 3D reconstruction and novel view synthesis (NVS) has rapidly evolved, particularly since the introduction of Neural Radiance Fields (NeRF)~\cite{mildenhall2020nerf}.
In this section, we review the most relevant works in the field, focusing on mesh and texture extraction methods.
All methods discussed below share the same input data: a set of multi-view images and camera poses.

\subsection{Multi-View Stereo}

Multi-view stereo (MVS) encompasses techniques that use stereo correspondences to reconstruct 3D geometry from multiple images~\cite{CGV-052}.
This approach has been extensively studied in computer vision for decades.
MVS methods can be broadly categorized into depth-map-based methods and volumetric-based methods.
Depth-map-based methods~\cite{Furukawa2010, Schonberger2016} estimate depth maps for each view and then fuse them into a coherent 3D model.
Volumetric methods~\cite{Kar2017, Seitz1999} represent the scene as a volumetric grid and iteratively refine it to match observed images.
While traditional MVS methods rely on handcrafted features and matching metrics, recent approaches leverage deep learning to improve depth estimation accuracy and robustness~\cite{yao2018mvsnet, ChenPMVSNet2019ICCV}.
Despite their success, MVS methods often struggle with textureless regions, reflective surfaces, and fine structures, resulting in incomplete or noisy reconstructions.

\subsection{Neural Radiance Field}

Mildenhall \etal~\cite{mildenhall2020nerf} introduced Neural Radiance Fields (NeRF) as a novel differentiable representation for 3D scenes.
NeRF represents scenes as continuous functions mapping 3D coordinates to radiance and volume density values, using neural networks as function approximators.
Barron \etal~\cite{barron2021mipnerf, barron2022mipnerf360} enhanced NeRF by introducing multi-scale representation and hierarchical sampling strategies.
They extended the method to unbounded scenes and reduced artifacts common in NeRF-based methods.
To extract meshes from NeRF representations, the marching cubes algorithm~\cite{William1987MarchingCubes} is typically applied to density fields evaluated on voxel grids.
However, due to network expressivity limitations and the fact that the model is not directly optimized for geometry capture, extracted meshes often appear coarse and lack fine details.

\subsection{Splatting-Based Methods}

Splatting-based methods represent scenes as sets of geometric primitives and employ a splatting technique introduced by Kerbl \etal~\cite{kerbl3Dgaussians} to render scenes efficiently and differentiably.
The original work~\cite{kerbl3Dgaussians} used 3D Gaussian primitives, while Huang \etal~\cite{Huang2DGS2024} proposed 2D Gaussian plates to enhance geometric accuracy.
Zhang \etal~\cite{zhang2024RaDe-GS} and Chen \etal~\cite{Chen2024PGSR} improved depth rendering by implementing more accurate depth calculation methods.
Hamdi \etal~\cite{hamdi_2024_CVPR} introduced GES to increase Gaussian kernel expressiveness while reducing the number of primitives needed for scene representation.
Held \etal~\cite{Held20243DConvex} further increased expressiveness by employing 3D convexes as primitives.
Other works~\cite{NEURIPS2024_ea13534e, zhang2024gspull} combine Gaussian Splatting with Signed Distance Functions (SDFs), jointly optimizing both representations while maintaining consistency between them.
These approaches constrain Gaussian primitives to lie on object surfaces within the scene.

For mesh extraction from splatting-based representations, the standard approach involves rendering depth maps for each view and fusing them using TSDF methods~\cite{Zhou2018, Richard2011KinectFusion}.
Since TSDF methods are grid-based, the resulting meshes cannot express varying levels of detail as effectively as the original primitive representation.

\subsection{Gradient-based Mesh Optimization}

Recent research has explored gradient-based optimization to directly reconstruct mesh and texture representations from multi-view images.
DMTet~\cite{shen2021dmtet} represents scenes as SDFs evaluated on deformable tetrahedral grids and implements differentiable marching tetrahedra for mesh extraction.
FlexiCubes~\cite{shen2023flexicubes} utilizes cubic grids and Dual Marching Cubes for mesh extraction, introducing additional parameters to enhance mesh representation flexibility.
While these approaches utilize 3D supervision, Munkberg \etal~\cite{Munkberg_2022_CVPR} built upon DMTet and employed deferred shading~\cite{Laine2020diffrast} to enable joint mesh and texture optimization with 2D image supervision.
MobileNeRF~\cite{chen2022mobilenerf} also optimizes colored meshes using differentiable rendering, with the mesh vertices initialized as a regular Euclidean lattice and locally optimized during training.
The mesh color is parameterized by a MLP and rendered using deferred shading.
However, being grid-based, these methods are constrained by grid resolution and cannot efficiently represent different levels of detail in a large-scale scene.

\subsection{Concurrent Work}

Differentiable rendering and 3D reconstruction remain active research areas with continuously emerging methods.
Here, we highlight recent works closely related to our approach.

Guo \etal~\cite{guo2025tetsphere} proposed modeling 3D shapes using deformable volumetric tetrahedral meshes.
Their method optimizes deformation matrices through gradient back-propagation from rendering errors and regularization terms constraining deformation energy.
Their Lagrangian representation avoids the iso-surface extraction required by Eulerian methods like DMTet and FlexiCubes and circumvents grid resolution limitations.
However, tetrahedral sphere representations are significantly more complex than 3D Gaussian representations, limiting applicability in large-scale scenes.

Tobiasz \etal~\cite{tobiasz2025meshsplats} developed a method to convert trained GS models to mesh models through post-processing.
They approximate Gaussian plates with multiple triangles and optimize triangle attributes using a differentiable renderer~\cite{Laine2020diffrast}.
While their motivation resembles ours, their extracted meshes contain several times more faces than the original Gaussian count, and the triangle faces remain transparent.
In contrast, our method maintains primitive count parity with Gaussian-based methods while producing solid meshes.

Besides, we have also noticed several works recently posted on arXiv that explore the use of triangle primitives in differentiable rendering in parallel to our approach~\cite{burgdorfer2025radianttrianglesoupsoft, held2025trianglesplattingrealtimeradiance, held2025trianglesplattingdifferentiablerendering}.
Our implementation provides a more accurate depth sorting method and also differs in the training pipeline compared to these works.
We acknowledge their contributions; however, our work was developed independently, and a detailed comparison is not provided at this stage.
\section{Method}
\label{sec:method}

\begin{figure}[t]
  \centering
  \includegraphics[width=0.99\columnwidth]{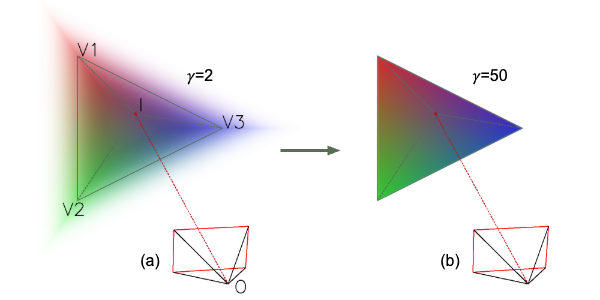}
  \caption{
      {\bf Overview of the proposed 2D Triangle Splatting (2DTS) method.}
      For each pixel on the rendered image, we calculate the intersection between the pixel ray and each triangle primitive in the 3D space.
      The opacity and color of each triangle's contribution to the pixel is dependent on the barycentric coordinates of the intersection point.
      The final color of the pixel is calculated by blending the colors of all triangles that cover the pixel.
      A compactness parameter $\mathbf{\gamma}$ is introduced to control the sharpness of the triangle edges.
      Panel (a) shows a triangle with $\mathbf{\gamma} = 2$, and panel (b) shows a triangle with $\mathbf{\gamma} = 50$.
  }
  \label{fig:overview}
\end{figure}

2D Triangle Splatting (2DTS) replaces the Gaussian primitives from 3DGS~\cite{kerbl3Dgaussians} with triangle primitives and combines the compactness parameter from GES~\cite{hamdi_2024_CVPR} to approximate a solid mesh representation.
The triangle primitives are rendered by alpha-blending the colors of all triangles intersecting each camera ray from front to back.
To enable the gradient propagation through rendering back to triangle vertices, we define the opacity of each point on the triangle plane as a continuous function of the barycentric coordinates.
The depth and normal of each primitive can be calculated naturally from the normal of the triangle plane and the location of ray-triangle intersection.
Depth maps and normal maps can then be rendered using the same alpha-blending method as the color rendering.
The parameters of the triangle primitives are optimized by minimizing a combination of photometric and regularization losses.
Figure \ref{fig:overview} shows an overview of the proposed method.

\subsection{Triangle Splatting}

Because we target triangle meshes that can be used directly in traditional rendering engines, our renderer should remain mathematically consistent with rasterization.
Starting from the rendering pipeline of 3DGS~\cite{kerbl3Dgaussians}, there are three main challenges to overcome.
First, the splatting process introduced in EWA~\cite{Zwicker2001EWA} and adopted by 3DGS uses a linear approximation of the perspective projection to project 3D Gaussians to the screen space.
We avoid this approximation by replacing splatting with ray-casting, which is mathematically exact.
Second, there is no straightforward way to define the depth of a Gaussian primitive in the camera space, and 3DGS sorts the Gaussians by their center's depth.
This approximation can lead to popping artifacts during camera rotation as discussed by Radl \etal~\cite{radl2024stopthepop}.
We adopt the method from their work with tile-based depth sorting and pixel-wise local resorting to achieve more accurate blending order.
Third, the primitives need to be partially transparent and diffuse to maintain differentiability, while traditional rasterization assumes solid surfaces.
To address this, we use Straight-Through Estimator (STE)~\cite{bengio2013} with compactness annealing to gradually convert diffuse triangle primitives into opaque triangle meshes during training.

\textbf{Ray-casting.} 
For a given pixel $\mathbf{P}_j$ on the image plane, we calculate the intersection point $\mathbf{I}_j$ between the ray shooting from the camera center $\mathbf{O}$ through $\mathbf{P}_j$ and each contributing triangle primitive.
Given the three vertices of the triangle $\mathbf{V}^1$, $\mathbf{V}^2$ and $\mathbf{V}^3$ and the pixel ray direction $\mathbf{r}_j$, $\mathbf{I}_j$ can be calculated by:
\begin{equation}
  \label{eq:ray_triangle_intersection}
  \begin{aligned}
    \mathbf{I}_j &= \mathbf{O} + d_j \cdot \mathbf{r}_j ,\\
    d_j &= \frac{(\mathbf{V}^1 - \mathbf{O}) \cdot \mathbf{N}}{\mathbf{r}_j \cdot \mathbf{N}} ,\\
    \mathbf{N} &= (\mathbf{V}^2 - \mathbf{V}^1) \times (\mathbf{V}^3 - \mathbf{V}^1) .
  \end{aligned}
\end{equation}
where $d_j$ and $\mathbf{N}$ are the depth and unnormalized normal respectively.

\textbf{Barycentric Coordinates.} 
With the ray-triangle intersection $\mathbf{I}_j$, we calculate the barycentric coordinates $(a^1_j, a^2_j, a^3_j)$ of the intersection point on the triangle plane.
\begin{equation}
  \label{eq:barycentric}
  \begin{aligned}
    a^1_j &= \frac{[(\mathbf{V}^2 - \mathbf{I}_j) \times (\mathbf{V}^3 - \mathbf{I}_j)] \cdot \mathbf{N}}{\mathbf{N} \cdot \mathbf{N}} ,\\
    a^2_j &= \frac{[(\mathbf{V}^3 - \mathbf{I}_j) \times (\mathbf{V}^1 - \mathbf{I}_j)] \cdot \mathbf{N}}{\mathbf{N} \cdot \mathbf{N}} ,\\
    a^3_j &= 1 - a^1_j - a^2_j .
  \end{aligned}
\end{equation}

\textbf{Color Interpolation.}
Gaussian-based representations always assume a constant color within each primitive, limiting their ability to represent high-frequency texture details.
With triangle primitives, we can naturally increase the expressiveness of each primitive using vertex color and interpolate the color at each pixel using barycentric coordinates:
\begin{equation}
  \label{eq:color}
  \mathbf{c}_j = a^1_j \cdot \mathbf{c}^1 + a^2_j \cdot \mathbf{c}^2 + a^3_j \cdot \mathbf{c}^3 ,
\end{equation}
where $\mathbf{c}^1$, $\mathbf{c}^2$ and $\mathbf{c}^3$ are the colors of the triangle vertices.
The vertex colors can either be native attributes of the triangle or evaluated from a set of spherical harmonic coefficients and the viewing direction, as introduced by Yu \etal~\cite{yu2022plenoxels}.

\textbf{Opacity Function.} 
To mimic the exponentially decaying opacity of the Gaussian primitives in 3DGS, we define the opacity of the triangle at each point $\mathbf{I}_j$ as:
\begin{equation}
  \label{eq:opacity}
  \begin{aligned}
    o_j &= O \cdot \exp(-\frac{1}{2} e_j^{2\gamma}) ,\\
    e_j &= 1 - 3 \cdot \min(a^1_j, a^2_j, a^3_j) .
  \end{aligned}
\end{equation}
where $O$ is the opacity property of the triangle, $\gamma$ is the compactness parameter we use to control the sharpness of the triangle edges, 
and $e_j$ is an eccentricity value analogous to the quadratic term in the exponent of a Gaussian kernel defined as:
\begin{equation}
  \mathcal{G}(\mathbf{x}) = \frac{1}{2\pi|\Sigma|^{\frac{1}{2}}} \exp(-\frac{1}{2} (\mathbf{x} - \mathbf{C})^T \Sigma (\mathbf{x} - \mathbf{C})) .
\end{equation}

\textbf{Opacity Annealing.} 
\label{sec:training_stages}
The eccentricity value $e_j$ defined above increases monotonically as the distance from $\mathbf{I}_j$ to the edge of the triangle decreases.
$e_j$ satisfies that $0 <= e_j < 1$ for points inside the triangle, $e_j = 1$ for points on the edge of the triangle, and $e_j > 1$ for points outside the triangle.
So, we have:
\begin{equation}
  \label{eq:opacity_limit}
  \begin{aligned}
    o_j|_{\gamma \to \infty, O \to 1} &=
    \begin{cases}
      1, & \mathbf{I}_j \quad \text{inside} ,\\
      0, & \mathbf{I}_j \quad \text{outside} . 
    \end{cases}
  \end{aligned}
\end{equation}
By gradually increasing the compactness parameter $\gamma$ and constraining $O$ to $1$, we can approximate a solid triangle mesh representation at the end of the training, as shown in Figure \ref{fig:overview}(b).
We maintain differentiability while ensuring a binary opacity $O$ during training by utilizing the Straight-Through Estimator (STE)~\cite{bengio2013}.
\begin{equation}
  \text{STE}(O) = (\text{H}(O - O_{\text{thres}}) - O).\text{detach}() + O ,
\end{equation}
where $\text{H}(\cdot)$ is the unit step function.
Triangles with opacity below $O_{\text{thres}}$ are pruned upon completion of training.

\textbf{Scale Compensation.}
During the increase of $\gamma$, the effective area covered by each triangle naturally shrinks. 
To maintain consistent coverage throughout the compactness tuning process, we implement a scale compensation mechanism based on an invariant opacity integration principle.

Mathematically, we aim to preserve the integration of opacity over the triangle plane regardless of the compactness parameter $\gamma$.
The integration can be calculated analytically as:
\begin{equation}
  \label{eq:scale-compensation}
  I(S, \gamma) = \int_{\triangle} o_j \, dA = S \cdot O \cdot \frac{2^{\frac{1}{\gamma}} \Gamma(\frac{1}{\gamma})}{\gamma} ,
\end{equation}
where $S$ is the triangle area, $\Gamma(x)$ is the Euler's Gamma function, and $o_j$ and $O$ are defined in Equation \ref{eq:opacity}.
The detailed derivation of this integration is provided in the supplementary materials.

As $\gamma$ approaches infinity, the integration becomes $I(S_0, \infty) = S_0 \cdot O$, with $S_0$ representing the solid triangle area. To ensure invariance across different $\gamma$ values, we apply a scale factor:
\begin{equation}
  S(\gamma) = S_0 \cdot \frac{\gamma}{2^{\frac{1}{\gamma}} \Gamma(\frac{1}{\gamma})} .
\end{equation}
This compensation ensures that triangles maintain their effective coverage throughout the optimization process, enabling smooth transition toward the solid mesh representation.

\textbf{Coverage Calculation.}
We follow 3DGS~\cite{kerbl3Dgaussians} and split the rendered image into tiles of $16\times 16$ pixels.
The coverage of each triangle on the tiles is determined by the bounding box of the triangle vertices projected onto the image plane.
Since the triangle's coverage is not limited to the area enclosed by its edges due to the continuous opacity function defined in Equation \ref{eq:opacity}, we expand the triangle by a scale factor $s = (2 \text{ln}(255))^{\frac{1}{2\gamma}}$ before projecting it to the image plane.

\textbf{Depth Sorting.}
\label{sec:depth_sorting}
The blending order of the triangles along each pixel ray is determined by the depth order of the intersection points calculated in Equation \ref{eq:ray_triangle_intersection}.
However, sorting the triangles for each pixel individually is computationally expensive.
To address this, we adopt a two-level depth sorting strategy similar to Radl \etal~\cite{radl2024stopthepop}.
First, the triangles are sorted tile-wise by the depth of the point of maximum opacity within each tile.
Triangles with maximum opacity below a threshold on each tile are discarded to reduce redundant computation.
Then, for each pixel, we perform a local re-sorting with a small depth buffer.
The depth buffer stores $K$ triangles along the pixel ray in depth order, and with each new triangle coming in, the closest triangle is blended and popped from the buffer.

\textbf{Alpha blending.}
With the triangles sorted and the opacity, depth, and normal of each triangle calculated for each pixel, we can render the final color, depth, and normal images through alpha blending from front to back as:
\begin{equation}
  \label{eq:alpha_blending}
  \left\{
  \begin{matrix}
    \mathbf{c}_j \\ 
    d_j \\ 
    \mathbf{N}_j
  \end{matrix}\right\} = 
  \sum_{i=1}^{N+1} \left\{
  \begin{matrix}
    \mathbf{c}^i_j \\ 
    d^i_j \\ 
    \frac{\mathbf{N}^i}{\|\mathbf{N}^i\|}
  \end{matrix}\right\} \cdot o^i_j T^i_j , \quad
  T^i_j = \prod_{k=1}^{i-1} (1 - o^k_j) .
\end{equation}
where $\mathbf{c}^i_j$ is defined in Equation \ref{eq:color}, $d^i_j$ and $\mathbf{N}^i$ are defined in Equation \ref{eq:ray_triangle_intersection}, and $o^i_j$ is defined in Equation \ref{eq:opacity}.
The superscript $i$ is the triangle index, and $i$ iterates over all $N$ triangles covering pixel $j$ in the sorted order.
In practice, we adopt an early termination strategy during the alpha-blending process when the accumulated transmittance $T^i_j$ drops below a certain threshold to reduce computation.
The blending also includes a background primitive with:
\begin{equation}
  \left\{
  \begin{matrix}
    \mathbf{c}^{N+1}_j \\ 
    d^{N+1}_j \\ 
    \mathbf{N}^{N+1}
  \end{matrix}\right\} = 
  \left\{
  \begin{matrix}
    \mathbf{c}_{bg} \\ 
    d_{bg} \\ 
    \mathbf{0}
  \end{matrix}\right\} .
\end{equation}

\subsection{Optimization}
\label{sec:optimization}

Because the rendered color, normal, and depth images are differentiable with respect to the triangle vertices, opacities, and colors, we can optimize the triangle attributes by minimizing a loss function defined on the rendered images.

\textbf{Geometric Constraints.} 
We use a normal consistency loss and a depth distortion loss similar to the ones used in 2DGS~\cite{Huang2DGS2024} to constrain the geometry of the triangles during training.
Given a rendered depth image $d_j$, we can infer a camera space normal image $\mathbf{n}'_j$ by calculating the depth gradient. 
We leave the details of the derivation to the supplementary materials.
With the rendered normal $\mathbf{n}_j$ and the inferred normal $\mathbf{n}'_j$, we define the normal consistency loss as:
\begin{equation}
  \mathcal{L}_{n} = \frac{1}{N} \sum_{j=1}^{N} \left( 1 - \mathbf{n}_j \cdot \mathbf{n}'_j \right) .
\end{equation}
The depth distortion loss is defined as:
\begin{equation}
  \mathcal{L}_{d} = \frac{1}{N} \sum_{j=1}^{N}\sum_{i=1}^{N}\sum_{k=1}^{i-1} o^i_j T^i_j o^k_j T^k_j (  d^i_j - d^k_j )^2 .
\end{equation}
where $i$ and $k$ are the indices of triangles covering pixel $j$.

\textbf{Loss Function.} 
We use an L1 loss and an SSIM loss to optimize the rendered color image $\mathbf{c}_j$ to the ground truth color image $\mathbf{c}_j^{gt}$.
The combined loss is defined as:
\begin{equation}
  \mathcal{L} = \lambda_{L1} \mathcal{L}_{L1} + \lambda_{SSIM} \mathcal{L}_{SSIM} + \lambda_n \mathcal{L}_n + \lambda_{d} \mathcal{L}_{d} ,
\end{equation}
where $\lambda_{L1}$, $\lambda_{SSIM}$, $\lambda_n$, and $\lambda_{d}$ are the weights of the corresponding loss terms.

\textbf{Densification.}
Because our reconstruction starts with a sparse point cloud from multi-view stereo methods, a densification process during training is necessary to fill in under-sampled areas.
We follow the densification strategy from Pixel-GS~\cite{zhang2024pixelgs} and calculate the sum of gradient norms on the vertices for each triangle.
Triangles with high gradient norms are split into four smaller triangles by adding mid-edge vertices.

\textbf{Contribution Pruning.}
In order to control the number of triangle primitives during the training, we calculate the sum and max contribution of each triangle to the final rendered image as:
\begin{equation}
  C^i_\text{sum} = \sum_{j=1}^{N} o^i_j T^i_j , \quad
  C^i_\text{max} = \max_{j=1}^{N} o^i_j T^i_j .
\end{equation}
where $i$ is the triangle index and $j$ is the pixel index. $o^i_j$ and $T^i_j$ are defined in Equation \ref{eq:opacity} and \ref{eq:alpha_blending} respectively.
Triangles with $C^i_\text{sum}$ or $C^i_\text{max}$ below a certain threshold are iteratively pruned during training.

\section{Experiments}
\label{sec:exp}

We conduct experiments of the following two categories.

In the first category, we optimize transparent and diffuse triangle primitives by fixing the compactness parameter $\gamma$ to 1 and letting opacities vary freely between 0 and 1.
The model in this setting is more expressive and can reconstruct the scene with higher visual quality.
We compare our method in this setting with other radiance field reconstruction methods on the Mip-NeRF360~\cite{barron2022mipnerf360}, NeRF-Synthetic~\cite{mildenhall2020nerf}, Tanks and Temples~\cite{Knapitsch2017TanksAndTemples}, and Deep Blending~\cite{Hedman2018DeepBlending} datasets.

\begin{figure}[ht]
    \centering
    \includegraphics[width=0.99\columnwidth]{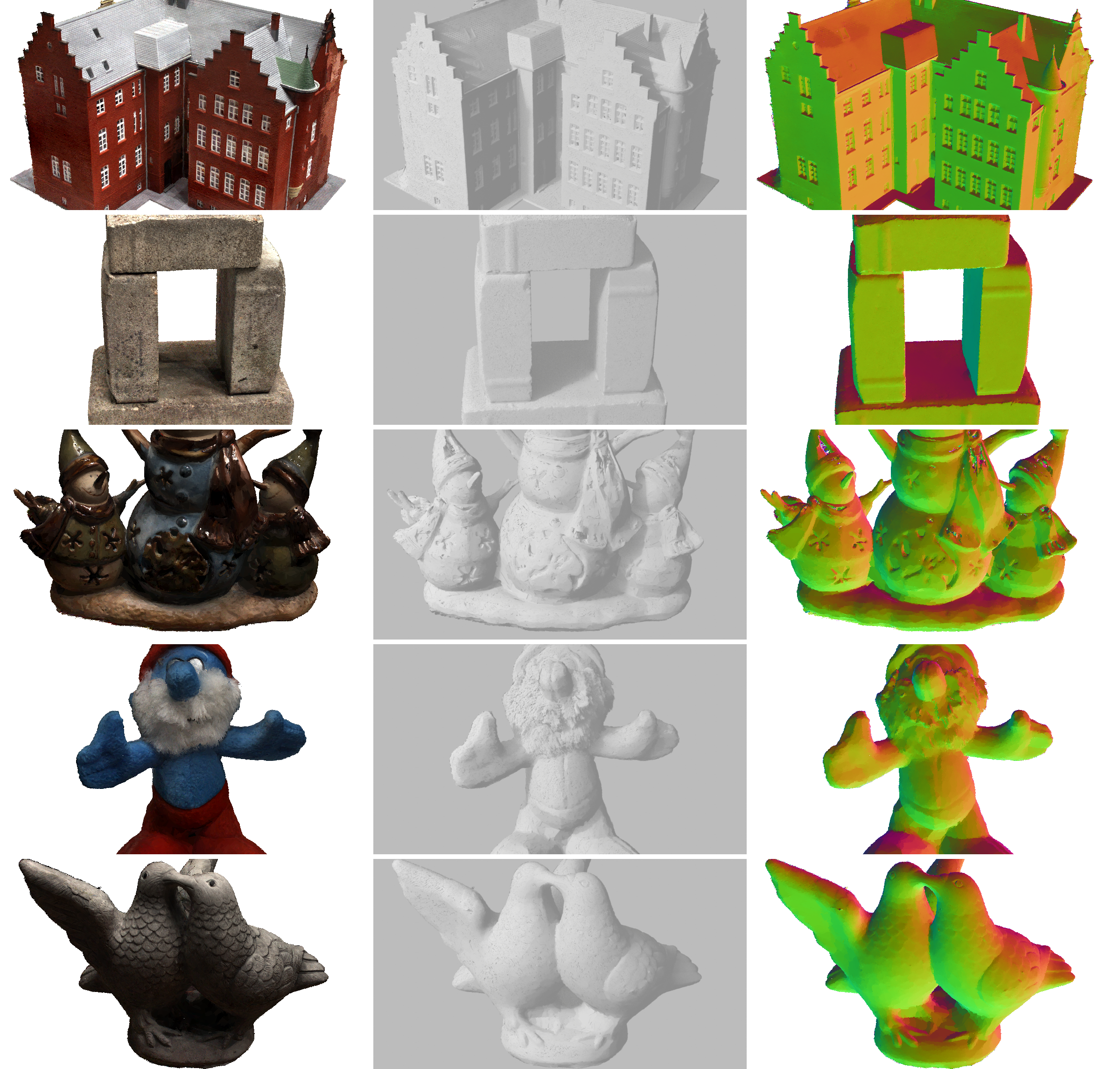}
    \caption{
        {\bf Demonstration of Meshes Reconstructed by 2DTS.}
        A visualization of meshes reconstructed by our method on the DTU~\cite{jensen2014large} dataset.
        From left to right: colored image rendered by our triangle splatting renderer, mesh rendered by Blender's Eevee engine, and normal map rendered by our renderer.
    }
    \label{fig:mesh-demo-dtu}
\end{figure}

In the second category, we optimize triangle meshes with opaque faces using binary opacities and a compactness parameter $\gamma$ gradually increased from 1 to 50 in an exponential schedule.
The model in this setting can better preserve the geometric structure of the scene with appropriate regularizations.
We compare our method in this setting with other mesh reconstruction methods on the NeRF-Synthetic~\cite{mildenhall2020nerf} and DTU~\cite{jensen2014large} datasets.
A visualization of the meshes reconstructed by our method on the DTU dataset is shown in Figure \ref{fig:mesh-demo-dtu}.

\subsection{Implementation}

We implement the 2D triangle splatting rasterizer with custom CUDA kernels.
For all experiments, we initialize the triangle primitives from a sparse point cloud generated by the structure from motion (SfM) pipeline of Colmap.
We set the barycenter and spherical harmonic coefficients of each triangle according to the location and color of the corresponding point.
Each triangle is initialized as an equilateral triangle with a scale proportional to the distance between the point and its nearest neighbor.
The normal of each triangle is initialized randomly.

We use the Adam optimizer to optimize the parameters of the triangle primitives.
The learning rates are modulated by exponential decay schedules.
Please refer to our code for the detailed learning rate settings.
All our experiments run on a single NVIDIA H20 GPU with 96 GB of memory.

\subsection{Experiments on Novel View Synthesis}
\label{sec:exp_cat1}

\begin{table*}[ht]
    \centering
    \small
    \resizebox{0.95\linewidth}{!}{
    \begin{tabular}{l|c c c|c c c|c c c|c c c}
                    & \multicolumn{3}{c|}{Mip-NeRF360}                                                  & \multicolumn{3}{c|}{NeRF-Synthetic}                                      & \multicolumn{3}{c|}{Tanks \& Temples}                                             & \multicolumn{3}{c}{Deep Blending}                                                 \\
                    & PSNR$\uparrow$            & SSIM$\uparrow$            & LPIPS$\downarrow$         & PSNR$\uparrow$         & SSIM$\uparrow$         & LPIPS$\downarrow$      & PSNR$\uparrow$            & SSIM$\uparrow$            & LPIPS$\downarrow$         & PSNR$\uparrow$            & SSIM$\uparrow$            & LPIPS$\downarrow$         \\
        \hline
        M-NeRF360   & \textcolor{2nd}{27.69}\td & 0.792\td                  & \textcolor{2nd}{0.237}\td & 33.05\ts{*}\td         & 0.961\ts{*}\td         & 0.067\ts{*}\td         & {22.22}\td                & {0.759}\td                & {0.267}\td                & 29.40\td                  & 0.901\td                  & \textcolor{2nd}{0.245}\td \\
        3DGS        & 27.24\td                  & {0.797}\td                & {0.246}\td                & \textcolor{2nd}{33.45} & \textcolor{1st}{0.970} & \textcolor{1st}{0.030} & \textcolor{2nd}{23.15}\td & {0.841}\td                & \textcolor{1st}{0.183}\td & \textcolor{2nd}{29.41}\td & \textcolor{2nd}{0.903}\td & \textcolor{1st}{0.243}\td \\
        2DGS        & 27.03\td                  & \textcolor{2nd}{0.804}\td & 0.239\td                  & 33.07                  & {0.968}                & \textcolor{2nd}{0.033} & 23.12                     & \textcolor{2nd}{0.842}    & {0.212}                   & \textcolor{1st}{29.55}    & \textcolor{2nd}{0.903}    & {0.257}                   \\
        2DTS (Ours) & \textcolor{1st}{28.18}    & \textcolor{1st}{0.842}    & \textcolor{1st}{0.218}    & \textcolor{1st}{33.51} & \textcolor{1st}{0.970} & 0.037                  & \textcolor{1st}{23.39}    & \textcolor{1st}{0.853}    & \textcolor{2nd}{0.204}    & {29.37}                   & \textcolor{1st}{0.908}    & {0.311}                   \\
    \end{tabular}
    }
    \caption{
        {\bf Comparison of Novel View Synthesis.} 
        Quantitative comparison of novel view synthesis of MipNeRF360, 3DGS, 2DGS and 2DTS on Mip-NeRF360~\cite{barron2022mipnerf360}, NeRF-Synthetic~\cite{mildenhall2020nerf}, Deep Blending~\cite{Hedman2018DeepBlending} and Tanks \& Temples~\cite{Knapitsch2017TanksAndTemples} datasets.
        For NeRF-Synthetic, we provide results from MipNeRF~\cite{barron2021mipnerf} instead of MipNeRF360, as marked by the \ts{*} sign, since NeRF-Synthetic includes bounded scenes.
        \td indicates results that are copied from the original papers.
        All other results are obtained by running the released official code by ourselves.
        \textcolor{1st}{Red} indicates the best score, \textcolor{2nd}{orange} indicates the second best score.
    }
    \label{tab:nvs}
\end{table*}

\begin{figure*}[ht]
    \centering
    \includegraphics[width=0.95\linewidth]{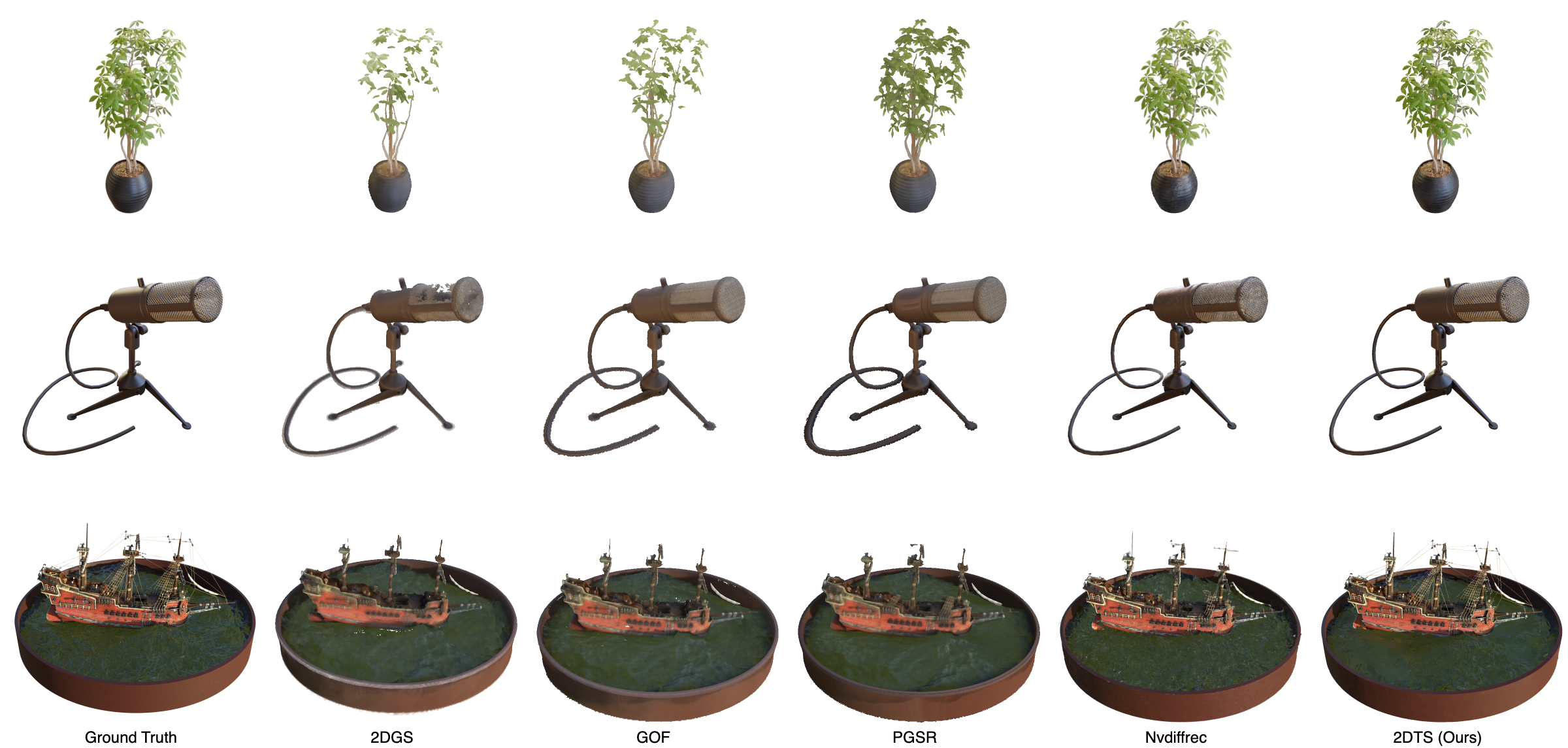}
    \caption{
        {\bf Comparison of Mesh Extraction.} 
        Visual comparison of meshes reconstructed by our method, 2DGS~\cite{Huang2DGS2024}, GOF~\cite{Yu2024GOF}, PGSR~\cite{Chen2024PGSR}, and Nvdiffrec~\cite{Munkberg_2022_CVPR} on the NeRF-Synthetic~\cite{mildenhall2020nerf} dataset.
        Our method better preserves finer structures such as the leaves of the ficus, the metal grille of the mic, and the ropes of the ship.
    }
    \label{fig:mesh-compare}
\end{figure*}

We first evaluate the first category against 3DGS~\cite{kerbl3Dgaussians}, 2DGS~\cite{Huang2DGS2024}, and MipNeRF~\cite{barron2021mipnerf}/MipNeRF360~\cite{barron2022mipnerf360} on Mip-NeRF360~\cite{barron2022mipnerf360}, NeRF-Synthetic~\cite{mildenhall2020nerf}, Tanks and Temples~\cite{Knapitsch2017TanksAndTemples}, and Deep Blending~\cite{Hedman2018DeepBlending}.
We use Peak Signal-to-Noise Ratio (PSNR), Structural Similarity Index Measure (SSIM) and Learned Perceptual Image Patch Similarity (LPIPS)~\cite{zhang2018perceptual} as evaluation metrics of novel view synthesis quality.
The quantitative results are shown in Table \ref{tab:nvs}.
Please refer to the supplementary materials for the detailed results on each scene.

All experiments of 2DTS in this setting run for 30k iterations with a batch size of 1.
The geometric regularization losses described in Section \ref{sec:optimization} are disabled in this category.
The maximum degree of spherical harmonic coefficients is set to 3.
Unless otherwise specified, all experiments of 2DTS in this setting use face color (uniform color per triangle).
For Mip-NeRF360, we use runtime bilinear downsampling, with indoor scene images downsampled by a factor of two and outdoor scene images downsampled by a factor of four. Images from other datasets are used as provided.
We create train/test splits by selecting every 8th photo as a test image for Mip-NeRF360, Tanks and Temples, and Deep Blending datasets, while using the provided split for NeRF-Synthetic.
All these configurations are kept consistent for all compared methods.

Our method achieves the best average PSNR and SSIM on Mip-NeRF360 and remains competitive on the other datasets, while LPIPS is mixed on some datasets.
We provide additional efficiency results, NVS ablations, and qualitative comparisons in the supplementary materials; see Tables~\ref{tab:rendering_speed_suppl} and \ref{tab:opacity_ablation_suppl}.

\subsection{Experiments on Mesh Reconstruction}
\label{sec:exp_cat2}

\begin{table*}[ht]
    \centering
    \resizebox{0.95\linewidth}{!}{
    \begin{tabular}{l|c c c c c c c|c c c c c c c}
                         & \multicolumn{7}{c|}{\makecell{NeRF-Synthetic}}                                                                                                                           & \multicolumn{7}{c}{\makecell{DTU}}                                                                                                                                           \\
                         & PSNR$\uparrow$         & SSIM$\uparrow$         & LPIPS$\downarrow$      & CD(1e-3)$\downarrow$  & Count(K)$\downarrow$ & Mem(GB)$\downarrow$    & Time(min)$\downarrow$ & PSNR$\uparrow$         & SSIM$\uparrow$          & LPIPS$\downarrow$      & CD$\downarrow$          & Count(K)$\downarrow$  & Mem(GB)$\downarrow$    & Time(min)$\downarrow$ \\
        \hline
        2DGS-Mesh        & 21.83                  & 0.866                  & 0.145                  & \textcolor{2nd}{47.5} & 489                  & \textcolor{1st}{0.31}  & {104}                 & 21.50                  & 0.851                   & 0.227                  & {0.819}                 & \textcolor{2nd}{268}  & \textcolor{1st}{0.55}  & {106}                 \\
        GOF-Mesh         & 22.13                  & 0.876                  & 0.136                  & {92.9}                & 501                  & {4.02}                 & {64}                  & 22.25                  & 0.869                   & 0.193                  & {0.812}                 & 1062                  & {4.74}                 & {92}                  \\
        PGSR-Mesh        & 21.71                  & 0.877                  & 0.134                  & {49.2}                & 361                  & {4.21}                 & \textcolor{2nd}{40}   & {22.73}                & \textcolor{2nd}{0.875}  & \textcolor{2nd}{0.190} & \textcolor{1st}{0.521}  & 1047                  & {1.91}                 & \textcolor{2nd} {40}  \\
        Nvdiffrec        & \textcolor{2nd}{28.76} & \textcolor{2nd}{0.938} & \textcolor{2nd}{0.080} & {72.5}                & \textcolor{2nd}{83}  & {2.12}                 & {85}                  & \textcolor{2nd}{25.89} & 0.860                   & 0.228                  & 1.907                   & 567                   & 7.07                   & {178}                 \\
        2DTS-Mesh (Ours) & \textcolor{1st}{30.80} & \textcolor{1st}{0.955} & \textcolor{1st}{0.066} & \textcolor{1st}{27.8} & \textcolor{1st}{73}  & \textcolor{2nd}{0.56}  & \textcolor{1st}{39}   & \textcolor{1st}{29.32} & \textcolor{1st}{0.944}  & \textcolor{1st}{0.145} & \textcolor{2nd}{0.570}  & \textcolor{1st}{265}  & \textcolor{2nd}{0.59}  & \textcolor{1st}{37}   \\
    \end{tabular}
    }
    \caption{
        {\bf Comparison of Mesh Extraction.} 
        Quantitative comparison of mesh reconstruction results of Nvdiffrec, 2DGS, GOF, PGSR and 2DTS on the NeRF-Synthetic~\cite{mildenhall2020nerf} and DTU~\cite{jensen2014large} datasets.
        Time only accounts for model training, while 2DGS, GOF and PGSR require additional time for mesh extraction using TSDF fusion.
        Memory only accounts for GPU usage (averaged reserved memory).
        CD stands for Chamfer Distance.
        Counts are the number of triangles in the extracted mesh.
        All values are averaged over all scenes.
        \textcolor{1st}{Red} indicates the best score, \textcolor{2nd}{orange} indicates the second best score.
    }
    \label{tab:mesh-compare}
\end{table*}

We then conduct experiments of the second category and compare our method with 2DGS~\cite{Huang2DGS2024}, Nvdiffrec~\cite{Munkberg_2022_CVPR}, GOF~\cite{Yu2024GOF} and PGSR~\cite{Chen2024PGSR} on the NeRF-Synthetic~\cite{mildenhall2020nerf} and DTU~\cite{jensen2014large} datasets.
The meshes from 2DGS, GOF and PGSR are extracted in a post-processing step using TSDF fusion~\cite{Zhou2018, Richard2011KinectFusion}, while our method and Nvdiffrec produce meshes directly from the optimization process.
The mesh from Nvdiffrec contains full PBR material and lighting information, while the meshes from other methods only contain vertex or face colors.
We use Chamfer Distance (CD) to compare the geometric accuracy of the reconstructed meshes.
We also compare the visual quality of the rendered images of the extracted meshes at test views using PSNR, SSIM and LPIPS metrics.
The meshes from Nvdiffrec and our method are rendered using the custom renderer, while the TSDF-extracted meshes from 2DGS, GOF and PGSR are rendered using the Kaolin library~\cite{KaolinLibrary}. 
The results are shown in Table \ref{tab:mesh-compare}, along with triangle face count, reconstruction time and GPU memory usage.
To reduce this renderer bias, we also report an apples-to-apples Kaolin comparison in the supplementary materials; see Table~\ref{tab:kaolin_rendering_suppl}.
A visual comparison of the colored meshes reconstructed by different methods is shown in Figure \ref{fig:mesh-compare}.
Please refer to the supplementary materials for the detailed results on each scene and visual comparisons on the DTU dataset.

All experiments of 2DTS in this setting run for 30K iterations with a batch size of 1.
In the first 20K iterations, we set the compactness parameter $\gamma$ to 1 to allow the triangle faces to remain diffuse.
In the last 10K iterations, we exponentially increase $\gamma$ from 1 to 50 to obtain triangle meshes with opaque faces.
An STE is applied to the triangle opacities from the beginning to the end of training.
The geometric regularization losses described in Section \ref{sec:optimization} are enabled in this setting to improve the geometric quality of the reconstructed meshes.
The maximum degree of spherical harmonic coefficients is kept at 3 during training, but only the zeroth order coefficients are retained in the extracted meshes when exported to the GLB format, so view-dependent appearance is not preserved in the standard mesh export.

The results for all compared methods are obtained by running the released official code ourselves with default settings.
The tetrahedral grid resolution of Nvdiffrec is set to 128 for NeRF-Synthetic and 256 for DTU.
The TSDF fusion voxel size for 2DGS, GOF and PGSR is set to 0.00787 for NeRF-Synthetic and 0.002 for DTU.

As demonstrated from the results, our method produces meshes with strong rendered visual quality and high geometric accuracy while maintaining much sparser triangles.
Notably, our method better preserves finer structures such as the leaves of the ficus, the metal grille of the mic, and the ropes of the ship, which are challenging for other methods to reconstruct.

\subsection{Ablation Study}
\label{sec:ablation}

To demonstrate the effectiveness of the geometric regularization losses introduced in Section \ref{sec:optimization} and the more accurate depth sorting introduced in Section \ref{sec:depth_sorting}, we conduct ablation studies on the DTU dataset.
We disable each component individually and compare the mesh reconstruction results with our full model.
The quantitative results are shown in Table \ref{tab:ablation}.

From the results, we can see that each proposed component contributes to the overall performance of our method.
Although disabling accurate depth sorting leads to slightly higher PSNR and SSIM scores, it results in worse Chamfer Distance scores, indicating that accurate depth sorting is beneficial for the geometric quality.

\begin{table}[ht]
    \centering
    \small
    \resizebox{0.99\columnwidth}{!}{
    \begin{tabular}{l|c c c c}
                                    & PSNR$\uparrow$         & SSIM$\uparrow$         & LPIPS$\downarrow$      & CD(1e-3)$\downarrow$   \\
        \hline
        Full Model                  & {29.32}                & {0.944}                & \textcolor{1st}{0.145} & \textcolor{1st}{0.570} \\
        w/o Normal Consistency Loss & 29.29                  & {0.939}                & 0.156                  & 0.686                  \\
        w/o Depth Distortion Loss   & 29.33                  & 0.944                  & \textcolor{1st}{0.145} & 0.575                  \\
        w/o Scale Compensation      & 29.11                  & 0.942                  & 0.149                  & 0.601                  \\
        w/o Accurate Depth Sorting  & \textcolor{1st}{29.57} & \textcolor{1st}{0.945} & \textcolor{1st}{0.145} & 0.580                  \\
    \end{tabular}
    }
    \caption{
        Ablation study on the DTU dataset.
        We disable each proposed component individually and compare the mesh reconstruction results with our full model.
        \textcolor{1st}{Red} indicates the best score.
    }
    \label{tab:ablation}
\end{table}

\section{Conclusion}
\label{sec:conclusion}

We present a novel method for mesh reconstruction from multi-view images using 2D triangle splatting.
This approach successfully combines the visual fidelity of radiance-field-based representations with the portability and efficiency of mesh-based representations.
Our experiments demonstrate strong rendered visual quality and geometric accuracy among the compared mesh reconstruction methods.
We emphasize that, to our knowledge, our method is the first to achieve differentiable mesh reconstruction without relying on any form of vertex grid.
This grid-free design is crucial for large-scale scene applications, enabling more flexible and efficient representation of varying detail levels.
We believe our method has significant potential for a wide range of applications, including 3D map reconstruction and virtual reality.
The code will be made publicly available.

\textbf{Limitations and Future Work.}
Although our method reconstructs meshes with opaque triangle faces and strong rendered quality, the result remains a disconnected triangle soup with possible overlaps and discontinuities.
As a result, the meshes produced by our method are not watertight, which makes them more suitable for rendering-oriented applications than simulation-related tasks.
There are two potential directions to address this limitation.
One is to apply a soft constraint by adding regularization terms that encourage neighboring triangles to share vertices during training.
Another direction is to directly use connected triangle meshes as primitives and accumulate the vertex gradients from all triangles sharing the same vertex.
These directions are left for future work.



{
    \small
    \bibliographystyle{ieeenat_fullname}
    \bibliography{main}

@String(CVPR= {IEEE Conf. Comput. Vis. Pattern Recog.})

@String(ICCV= {Int. Conf. Comput. Vis.})

@String(ECCV= {Eur. Conf. Comput. Vis.})

@String(TOG= {ACM Trans. Graph.})

@String(CVPR  = {CVPR})

@String(ICCV  = {ICCV})

@String(ECCV  = {ECCV})

@String(TOG   = {ACM TOG})

@article{kerbl3Dgaussians,
  author  = {Kerbl, Bernhard and Kopanas, Georgios and Leimk{\"u}hler, Thomas and Drettakis, George},
  title   = {{3D} Gaussian Splatting for Real-Time Radiance Field Rendering},
  journal = {ACM Transactions on Graphics},
  number  = {4},
  volume  = {42},
  month   = {July},
  year    = {2023},
  url     = {https://repo-sam.inria.fr/fungraph/3d-gaussian-splatting/}
}

@article{Yu2023MipSplatting,
  author  = {Yu, Zehao and Chen, Anpei and Huang, Binbin and Sattler, Torsten and Geiger, Andreas},
  title   = {{Mip-Splatting}: Alias-free {3D} Gaussian Splatting},
  journal = {Conference on Computer Vision and Pattern Recognition (CVPR)},
  year    = {2024}
}

@inproceedings{hamdi_2024_CVPR,
  title     = {{GES}: Generalized Exponential Splatting for Efficient Radiance Field Rendering},
  author    = {Hamdi, Abdullah and Melas-Kyriazi, Luke and Mai, Jinjie and Qian, Guocheng and Liu, Ruoshi and Vondrick, Carl and Ghanem, Bernard and Vedaldi, Andrea},
  booktitle = {Proceedings of the IEEE/CVF Conference on Computer Vision and Pattern Recognition (CVPR)},
  month     = {June},
  year      = {2024},
  pages     = {19812-19822}
}

@inproceedings{guo2025tetsphere,
  title     = {{TetSphere Splatting}: Representing High-Quality Geometry with {Lagrangian} Volumetric Meshes},
  author    = {Minghao Guo and Bohan Wang and Kaiming He and Wojciech Matusik},
  booktitle = {The Thirteenth International Conference on Learning Representations},
  year      = {2025},
  url       = {https://openreview.net/forum?id=8enWnd6Gp3}
}

@inproceedings{Huang2DGS2024,
  title     = {{2D} Gaussian Splatting for Geometrically Accurate Radiance Fields},
  author    = {Huang, Binbin and Yu, Zehao and Chen, Anpei and Geiger, Andreas and Gao, Shenghua},
  publisher = {Association for Computing Machinery},
  booktitle = {SIGGRAPH 2024 Conference Papers},
  year      = {2024},
  doi       = {10.1145/3641519.3657428}
}

@article{Held20243DConvex,
  title      = {{3D} Convex Splatting: Radiance Field Rendering with {3D} Smooth Convexes},
  author     = {Held, Jan and Vandeghen, Renaud and Hamdi, Abdullah and Deli{`e}ge, Adrien and Cioppa, Anthony and Giancola, Silvio and Vedaldi, Andrea and Ghanem, Bernard and Van Droogenbroeck, Marc},
  journal    = {arXiv},
  volume     = {abs/2411.14974},
  year       = {2024},
  eprint     = {2411.14974},
  eprinttype = {arXiv},
  doi        = {10.48550/arXiv.2411.14974},
  url        = {https://doi.org/10.48550/arXiv.2411.14974}
}

@article{Chen2024PGSR,
  author  = {Chen, Danpeng and Li, Hai and Ye, Weicai and Wang, Yifan and Xie, Weijian and Zhai, Shangjin and Wang, Nan and Liu, Haomin and Bao, Hujun and Zhang, Guofeng},
  journal = {IEEE Transactions on Visualization and Computer Graphics},
  title   = {{PGSR}: Planar-based Gaussian Splatting for Efficient and High-Fidelity Surface Reconstruction},
  year    = {2024},
  pages   = {1-12},
  doi     = {10.1109/TVCG.2024.3494046}
}

@inproceedings{mildenhall2020nerf,
  title     = {{NeRF}: Representing Scenes as Neural Radiance Fields for View Synthesis},
  author    = {Ben Mildenhall and Pratul P. Srinivasan and Matthew Tancik and Jonathan T. Barron and Ravi Ramamoorthi and Ren Ng},
  year      = {2020},
  booktitle = {ECCV}
}

@misc{lin2024metasapiens,
  title         = {{MetaSapiens}: Real-Time Neural Rendering with Efficiency-Aware Pruning and Accelerated Foveated Rendering},
  author        = {Weikai Lin and Yu Feng and Yuhao Zhu},
  year          = {2024},
  eprint        = {2407.00435},
  archiveprefix = {arXiv},
  primaryclass  = {cs.GR},
  doi           = {https://doi.org/10.1145/3669940.3707227},
  url           = {https://arxiv.org/abs/2407.00435}
}

@misc{zhang2024RaDe-GS,
  title         = {{RaDe-GS}: Rasterizing Depth in Gaussian Splatting},
  author        = {Baowen Zhang and Chuan Fang and Rakesh Shrestha and Yixun Liang and Xiaoxiao Long and Ping Tan},
  year          = {2024},
  eprint        = {2406.01467},
  archiveprefix = {arXiv},
  primaryclass  = {cs.GR},
  url           = {https://arxiv.org/abs/2406.01467}
}

@inproceedings{Richard2011KinectFusion,
  author    = {Newcombe, Richard A. and Izadi, Shahram and Hilliges, Otmar and Molyneaux, David and Kim, David and Davison, Andrew J. and Kohi, Pushmeet and Shotton, Jamie and Hodges, Steve and Fitzgibbon, Andrew},
  booktitle = {2011 10th IEEE International Symposium on Mixed and Augmented Reality},
  title     = {{KinectFusion}: Real-time dense surface mapping and tracking},
  year      = {2011},
  volume    = {},
  number    = {},
  pages     = {127-136},
  doi       = {10.1109/ISMAR.2011.6092378}
}

@article{Zhou2018,
  author  = {Qian-Yi Zhou and Jaesik Park and Vladlen Koltun},
  title   = {{Open3D}: {A} Modern Library for {3D} Data Processing},
  journal = {arXiv:1801.09847},
  year    = {2018}
}

@inproceedings{Gao2024Relightable,
  author    = {Gao, Jian and Gu, Chun and Lin, Youtian and Li, Zhihao and Zhu, Hao and Cao, Xun and Zhang, Li and Yao, Yao},
  title     = {{Relightable 3D Gaussians}: Realistic Point Cloud Relighting with {BRDF} Decomposition and Ray Tracing},
  year      = {2024},
  isbn      = {978-3-031-72994-2},
  publisher = {Springer-Verlag},
  address   = {Berlin, Heidelberg},
  url       = {https://doi.org/10.1007/978-3-031-72995-9_5},
  doi       = {10.1007/978-3-031-72995-9_5},
  booktitle = {ECCV},
  pages     = {73-89},
  numpages  = {17},
  location  = {Milan, Italy}
}

@article{shen2023flexicubes,
  author     = {Shen, Tianchang and Munkberg, Jacob and Hasselgren, Jon and Yin, Kangxue and Wang, Zian 
                and Chen, Wenzheng and Gojcic, Zan and Fidler, Sanja and Sharp, Nicholas and Gao, Jun},
  title      = {Flexible Isosurface Extraction for Gradient-Based Mesh Optimization},
  year       = {2023},
  issue_date = {August 2023},
  publisher  = {Association for Computing Machinery},
  address    = {New York, NY, USA},
  volume     = {42},
  number     = {4},
  issn       = {0730-0301},
  url        = {https://doi.org/10.1145/3592430},
  doi        = {10.1145/3592430},
  journal    = {ACM Trans. Graph.},
  month      = {jul},
  articleno  = {37},
  numpages   = {16}
}

@inproceedings{shen2021dmtet,
  title     = {Deep Marching Tetrahedra: a Hybrid Representation for High-Resolution 3D Shape Synthesis},
  author    = {Tianchang Shen and Jun Gao and Kangxue Yin and Ming-Yu Liu and Sanja Fidler},
  year      = {2021},
  booktitle = {Advances in Neural Information Processing Systems (NeurIPS)}
}

@inproceedings{Munkberg_2022_CVPR,
  author    = {Munkberg, Jacob and Hasselgren, Jon and Shen, Tianchang and Gao, Jun and Chen, Wenzheng 
               and Evans, Alex and M\"uller, Thomas and Fidler, Sanja},
  title     = {{Extracting Triangular 3D Models, Materials, and Lighting From Images}},
  booktitle = {Proceedings of the IEEE/CVF Conference on Computer Vision and Pattern Recognition (CVPR)},
  month     = {June},
  year      = {2022},
  pages     = {8280-8290}
}

@article{CGV-052,
  url     = {http://dx.doi.org/10.1561/0600000052},
  year    = {2015},
  volume  = {9},
  journal = {Foundations and Trends® in Computer Graphics and Vision},
  title   = {Multi-View Stereo: A Tutorial},
  doi     = {10.1561/0600000052},
  issn    = {1572-2740},
  number  = {1-2},
  pages   = {1-148},
  author  = {Yasutaka Furukawa and Carlos Hernández}
}

@article{William1987MarchingCubes,
  author     = {Lorensen, William E. and Cline, Harvey E.},
  title      = {Marching cubes: A high resolution 3D surface construction algorithm},
  year       = {1987},
  issue_date = {July 1987},
  publisher  = {Association for Computing Machinery},
  address    = {New York, NY, USA},
  volume     = {21},
  number     = {4},
  issn       = {0097-8930},
  url        = {https://doi.org/10.1145/37402.37422},
  doi        = {10.1145/37402.37422},
  journal    = {SIGGRAPH Comput. Graph.},
  month      = aug,
  pages      = {163-169},
  numpages   = {7}
}

@inproceedings{zhang2024gspull,
  title     = {Neural Signed Distance Function Inference through Splatting 3D Gaussians Pulled on Zero-Level Set},
  author    = {Wenyuan Zhang and Yu-Shen Liu and Zhizhong Han},
  booktitle = {Advances in Neural Information Processing Systems},
  year      = {2024}
}

@misc{barron2021mipnerf,
  title         = {{Mip-NeRF}: A Multiscale Representation for Anti-Aliasing Neural Radiance Fields},
  author        = {Jonathan T. Barron and Ben Mildenhall and Matthew Tancik and Peter Hedman and Ricardo Martin-Brualla and Pratul P. Srinivasan},
  year          = {2021},
  eprint        = {2103.13415},
  archiveprefix = {arXiv},
  primaryclass  = {cs.CV}
}

@article{barron2022mipnerf360,
  title   = {{Mip-NeRF 360}: Unbounded Anti-Aliased Neural Radiance Fields},
  author  = {Jonathan T. Barron and Ben Mildenhall and Dor Verbin and Pratul P. Srinivasan and Peter Hedman},
  journal = {CVPR},
  year    = {2022}
}

@inproceedings{NEURIPS2024_ea13534e,
  author    = {Yu, Mulin and Lu, Tao and Xu, Linning and Jiang, Lihan and Xiangli, Yuanbo and Dai, Bo},
  booktitle = {Advances in Neural Information Processing Systems},
  editor    = {A. Globerson and L. Mackey and D. Belgrave and A. Fan and U. Paquet and J. Tomczak and C. Zhang},
  pages     = {129507--129530},
  publisher = {Curran Associates, Inc.},
  title     = {{GSDF}: {3DGS} Meets {SDF} for Improved Neural Rendering and Reconstruction},
  url       = {https://proceedings.neurips.cc/paper_files/paper/2024/file/ea13534ee239bb3977795b8cc855bacc-Paper-Conference.pdf},
  volume    = {37},
  year      = {2024}
}

@article{Laine2020diffrast,
  title   = {Modular Primitives for High-Performance Differentiable Rendering},
  author  = {Samuli Laine and Janne Hellsten and Tero Karras and Yeongho Seol and Jaakko Lehtinen and Timo Aila},
  journal = {ACM Transactions on Graphics},
  year    = {2020},
  volume  = {39},
  number  = {6}
}

@misc{tobiasz2025meshsplats,
  title         = {{MeshSplats}: Mesh-Based Rendering with Gaussian Splatting Initialization},
  author        = {Rafał Tobiasz and Grzegorz Wilczyński and Marcin Mazur and Sławomir Tadeja and Przemysław Spurek},
  year          = {2025},
  eprint        = {2502.07754},
  archiveprefix = {arXiv},
  primaryclass  = {cs.GR},
  url           = {https://arxiv.org/abs/2502.07754}
}

@inproceedings{Zwicker2001EWA,
  author    = {Zwicker, M. and Pfister, H. and van Baar, J. and Gross, M.},
  booktitle = {Proceedings Visualization, 2001. VIS '01.},
  title     = {EWA volume splatting},
  year      = {2001},
  volume    = {},
  number    = {},
  pages     = {29-538},
  doi       = {10.1109/VISUAL.2001.964490}
}

@inproceedings{yu2022plenoxels,
  title     = {Plenoxels: Radiance Fields without Neural Networks},
  author    = {Sara Fridovich-Keil and Alex Yu and Matthew Tancik and Qinhong Chen and Benjamin Recht and Angjoo Kanazawa},
  year      = {2022},
  booktitle = {CVPR}
}

@article{Furukawa2010,
  author   = {Furukawa, Yasutaka and Ponce, Jean},
  journal  = {IEEE Transactions on Pattern Analysis and Machine Intelligence},
  title    = {Accurate, Dense, and Robust Multiview Stereopsis},
  year     = {2010},
  volume   = {32},
  number   = {8},
  pages    = {1362-1376},
  doi      = {10.1109/TPAMI.2009.161}
}

@inproceedings{Schonberger2016,
  author    = {Sch{\"o}nberger, Johannes L. and Zheng, Enliang and Frahm, Jan-Michael and Pollefeys, Marc},
  editor    = {Leibe, Bastian and Matas, Jiri and Sebe, Nicu and Welling, Max},
  title     = {Pixelwise View Selection for Unstructured Multi-View Stereo},
  booktitle = {Computer Vision -- ECCV 2016},
  year      = {2016},
  publisher = {Springer International Publishing},
  address   = {Cham},
  pages     = {501--518},
  isbn      = {978-3-319-46487-9}
}

@article{yao2018mvsnet,
  title   = {MVSNet: Depth Inference for Unstructured Multi-view Stereo},
  author  = {Yao, Yao and Luo, Zixin and Li, Shiwei and Fang, Tian and Quan, Long},
  journal = {European Conference on Computer Vision (ECCV)},
  year    = {2018}
}

@inproceedings{ChenPMVSNet2019ICCV,
  author    = {Chen, Rui and Han, Songfang and Xu, Jing and Su, Hao},
  title     = {Point-based Multi-view Stereo Network},
  booktitle = {The IEEE International Conference on Computer Vision (ICCV)},
  year      = {2019}
}

@inproceedings{Seitz1999,
  author    = {Kutulakos, K.N. and Seitz, S.M.},
  booktitle = {Proceedings of the Seventh IEEE International Conference on Computer Vision},
  title     = {A theory of shape by space carving},
  year      = {1999},
  volume    = {1},
  number    = {},
  pages     = {307-314 vol.1},
  doi       = {10.1109/ICCV.1999.791235}
}

@inproceedings{Kar2017,
  author    = {Kar, Abhishek and H\"{a}ne, Christian and Malik, Jitendra},
  booktitle = {Advances in Neural Information Processing Systems},
  editor    = {I. Guyon and U. Von Luxburg and S. Bengio and H. Wallach and R. Fergus and S. Vishwanathan and R. Garnett},
  pages     = {},
  publisher = {Curran Associates, Inc.},
  title     = {Learning a Multi-View Stereo Machine},
  url       = {https://proceedings.neurips.cc/paper_files/paper/2017/file/9c838d2e45b2ad1094d42f4ef36764f6-Paper.pdf},
  volume    = {30},
  year      = {2017}
}

@article{Knapitsch2017TanksAndTemples,
  author    = {Arno Knapitsch and Jaesik Park and Qian-Yi Zhou and Vladlen Koltun},
  title     = {Tanks and Temples: Benchmarking Large-Scale Scene Reconstruction},
  journal   = {ACM Transactions on Graphics (ToG)},
  volume    = {36},
  number    = {4},
  pages     = {1--13},
  year      = {2017},
  publisher = {ACM},
  address   = {New York, NY, USA},
  doi       = {10.1145/3072959.3073693},
  note      = {6, 11, 13, 14}
}

@article{Hedman2018DeepBlending,
  author    = {Peter Hedman and Julien Philip and True Price and Jan-Michael Frahm and George Drettakis and Gabriel Brostow},
  title     = {Deep Blending for Free-viewpoint Image-based Rendering},
  journal   = {ACM Transactions on Graphics (Proc. SIGGRAPH Asia)},
  volume    = {37},
  number    = {6},
  pages     = {257:1--257:15},
  year      = {2018},
  publisher = {ACM},
  doi       = {10.1145/3272127.3275079},
  note      = {6, 11, 13, 14}
}

@software{KaolinLibrary,
  author  = {Fuji Tsang, Clement and Shugrina, Maria and Lafleche, Jean Francois and Perel, Or and Loop, Charles and Takikawa, Towaki and Modi, Vismay and Zook, Alexander and Wang, Jiehan and Chen, Wenzheng and Shen, Tianchang and Gao, Jun and Jatavallabhula, Krishna Murthy and Smith, Edward and Rozantsev, Artem and Fidler, Sanja and State, Gavriel and Gorski, Jason and Xiang, Tommy and Li, Jianing and Li, Michael and Lebaredian, Rev},
  title   = {Kaolin: A Pytorch Library for Accelerating 3D Deep Learning Research},
  date    = {2024-11-20},
  version = {0.17.0},
  url     = {\url{https://github.com/NVIDIAGameWorks/kaolin}}
}

@inproceedings{chen2022mobilenerf,
  title     = {{MobileNeRF}: Exploiting the Polygon Rasterization Pipeline for Efficient Neural Field Rendering on Mobile Architectures},
  author    = {Zhiqin Chen and Thomas Funkhouser and Peter Hedman and Andrea Tagliasacchi},
  booktitle = {The Conference on Computer Vision and Pattern Recognition (CVPR)},
  year      = {2023}
}

@misc{burgdorfer2025radianttrianglesoupsoft,
  title         = {Radiant Triangle Soup with Soft Connectivity Forces for 3D Reconstruction and Novel View Synthesis},
  author        = {Nathaniel Burgdorfer and Philippos Mordohai},
  year          = {2025},
  eprint        = {2505.23642},
  archiveprefix = {arXiv},
  primaryclass  = {cs.CV},
  url           = {https://arxiv.org/abs/2505.23642}
}

@misc{held2025trianglesplattingrealtimeradiance,
  title         = {Triangle Splatting for Real-Time Radiance Field Rendering},
  author        = {Jan Held and Renaud Vandeghen and Adrien Deliege and Abdullah Hamdi and Silvio Giancola and Anthony Cioppa and Andrea Vedaldi and Bernard Ghanem and Andrea Tagliasacchi and Marc Van Droogenbroeck},
  year          = {2025},
  eprint        = {2505.19175},
  archiveprefix = {arXiv},
  primaryclass  = {cs.CV},
  url           = {https://arxiv.org/abs/2505.19175}
}

@misc{held2025trianglesplattingdifferentiablerendering,
  title         = {Triangle Splatting+: Differentiable Rendering with Opaque Triangles},
  author        = {Jan Held and Renaud Vandeghen and Sanghyun Son and Daniel Rebain and Matheus Gadelha and Yi Zhou and Ming C. Lin and Marc Van Droogenbroeck and Andrea Tagliasacchi},
  year          = {2025},
  eprint        = {2509.25122},
  archiveprefix = {arXiv},
  primaryclass  = {cs.CV},
  url           = {https://arxiv.org/abs/2509.25122}
}

@article{radl2024stopthepop,
  author    = {Radl, Lukas and Steiner, Michael and Parger, Mathias and Weinrauch, Alexander and Kerbl, Bernhard and Steinberger, Markus},
  title     = {{StopThePop: Sorted Gaussian Splatting for View-Consistent Real-time Rendering}},
  journal   = {ACM Transactions on Graphics},
  number    = {4},
  volume    = {43},
  articleno = {64},
  year      = {2024}
}

@misc{bengio2013,
  title         = {Estimating or Propagating Gradients Through Stochastic Neurons for Conditional Computation},
  author        = {Yoshua Bengio and Nicholas Léonard and Aaron Courville},
  year          = {2013},
  eprint        = {1308.3432},
  archiveprefix = {arXiv},
  primaryclass  = {cs.LG},
  url           = {https://arxiv.org/abs/1308.3432}
}

@inproceedings{zhang2024pixelgs,
  title     = {Pixel-GS: Density Control with Pixel-aware Gradient for 3D Gaussian Splatting},
  author    = {Zhang, Zheng and Hu, Wenbo and Lao, Yixing and He, Tong and Zhao, Hengshuang},
  booktitle = {ECCV},
  year      = {2024}
}

@inproceedings{jensen2014large,
  title        = {Large scale multi-view stereopsis evaluation},
  author       = {Jensen, Rasmus and Dahl, Anders and Vogiatzis, George and Tola, Engil and Aan{\ae}s, Henrik},
  booktitle    = {2014 IEEE Conference on Computer Vision and Pattern Recognition},
  pages        = {406--413},
  year         = {2014},
  organization = {IEEE}
}

@article{Yu2024GOF,
  author  = {Yu, Zehao and Sattler, Torsten and Geiger, Andreas},
  title   = {Gaussian Opacity Fields: Efficient Adaptive Surface Reconstruction in Unbounded Scenes},
  journal = {ACM Transactions on Graphics},
  year    = {2024}
}

@inproceedings{zhang2018perceptual,
  title     = {The Unreasonable Effectiveness of Deep Features as a Perceptual Metric},
  author    = {Zhang, Richard and Isola, Phillip and Efros, Alexei A and Shechtman, Eli and Wang, Oliver},
  booktitle = {CVPR},
  year      = {2018}
}
}

\clearpage
\setcounter{page}{1}
\maketitlesupplementary

\section{Detailed Proof of Equations}

\subsection{Render Normal from Depth}

Given a rendered depth image $d_j$, we can infer a camera space normal image $\mathbf{n}'_j$ as follows.
We first define:
\begin{equation}
    \begin{aligned}
      \mathbf{D}_j &= \left[ D_{x}, D_{y} \right]\vert_{\mathbf{P}_j} = \left[ \frac{1}{d} \frac{\partial d}{\partial x}, \frac{1}{d} \frac{\partial d}{\partial y} \right]\vert_{\mathbf{P}_j} ,\\
      \mathbf{n}'_j &= \text{norm} \left( \left[ n'_{x}, n'_{y}, n'_{z} \right]\vert_{\mathbf{P}_j} \right) ,
    \end{aligned}
\end{equation}
where $\frac{\partial d}{\partial x}$ and $\frac{\partial d}{\partial y}$ are the depth gradient on the pixel $\mathbf{P}_j$ defined as the difference between the depth of adjacent pixels, which we calculate with a Scharr Filter.
Then, we have:
\begin{equation}
    \begin{aligned}
        n'_{x} &= \frac{WD_{x}}{2~\text{tan}(\frac{\text{FoV}_x}{2})} ,\\
        n'_{y} &= \frac{HD_{y}}{2~\text{tan}(\frac{\text{FoV}_y}{2})} ,\\
        n'_{z} &= 1 + D_{x}(x - \frac{W}{2} + 0.5) + D_{y}(y - \frac{H}{2} + 0.5) ,\\
    \end{aligned}
\end{equation}
where $W$ and $H$ are the width and height of the image, $x$ and $y$ are the pixel coordinates.

\subsection{Integration of Opacity}

\begin{figure}[ht]
    \centering
    \includegraphics[width=0.99\columnwidth]{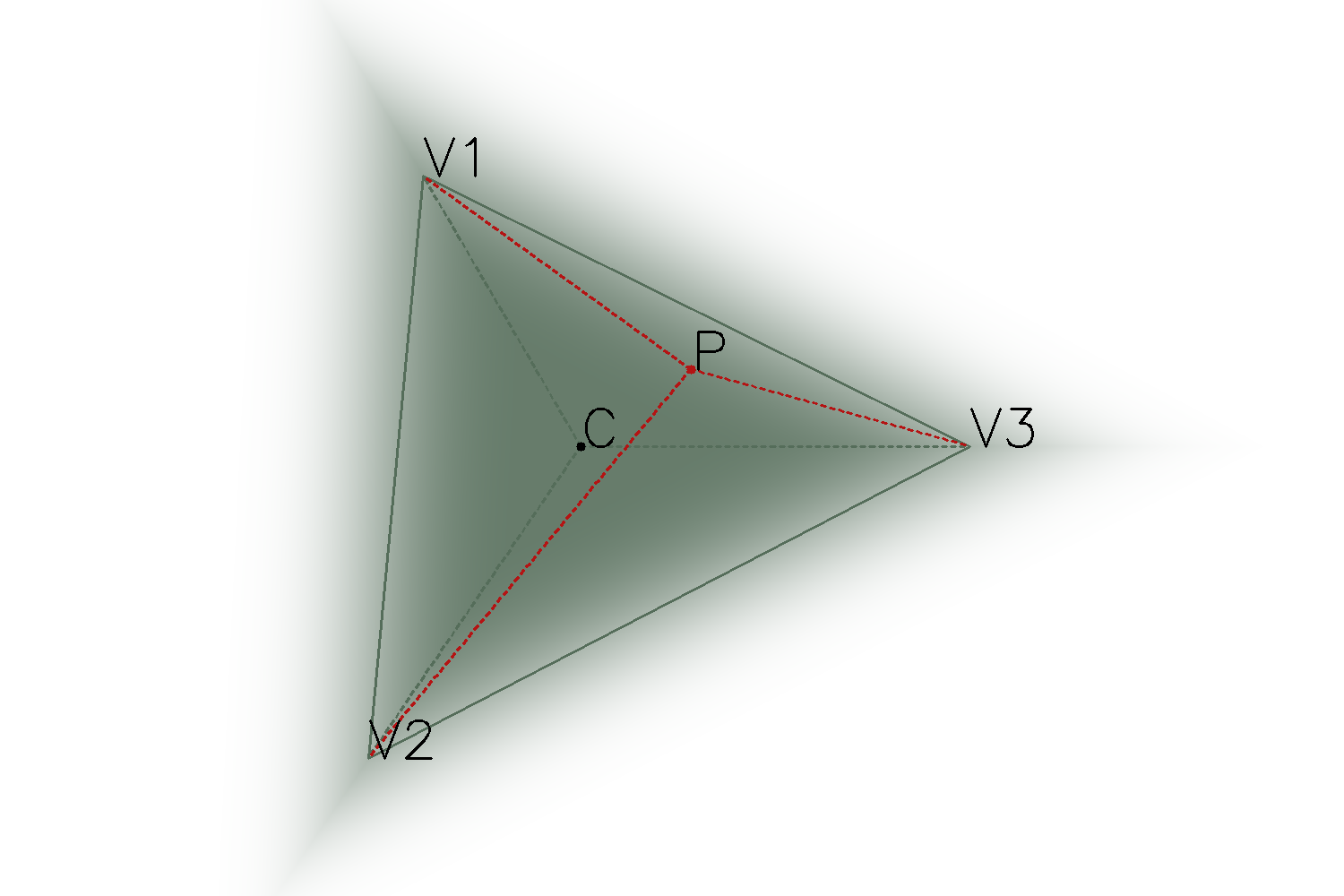}
    \caption{
        {\bf Auxiliary Chart for the Integration of Opacity.}
    }
    \label{fig:chart}
\end{figure}

We provide the detailed derivation of the integration of opacity over the triangle plane used in Equation \ref{eq:scale-compensation} of the main paper.

\begin{equation}
    \begin{aligned}
        I(\gamma) &= \int_{\triangle} o_j \, dA = \int_{\triangle} O \cdot \exp(-\frac{1}{2} e_j^{2\gamma}) \, dA \\
    \end{aligned}
\end{equation}

We calculate the integration separately for the three subregions of the triangle divided by $\overline{CV_1}$, $\overline{CV_2}$, and $\overline{CV_3}$, where $C$ is the barycenter, as shown in Figure \ref{fig:chart}.
For the subregion enclosed by $\overline{CV_1}$ and $\overline{CV_2}$, we can transform the integration over area to integration over distance $h$ to the edge $\overline{V_1V_2}$ since the opacity is constant at each point on a line parallel to $\overline{V_1V_2}$ in this subregion.
So we have:

\begin{equation}
    \begin{aligned}
        &I_{\overline{CV_1}\vee\overline{CV_2}}(\gamma) = \int_{-\infty}^{H} O \cdot \exp(-\frac{1}{2} e_{(h)}^{2\gamma}) \frac{H-h}{H} L_{\overline{V_1V_2}} dh \\
        &= \int_{-\infty}^{H} O \cdot \exp(-\frac{1}{2} (1-\frac{h}{H})^{2\gamma}) (1 - \frac{h}{H}) L_{\overline{V_1V_2}} dh \\
        &= \int_{0}^{\infty} \frac{1}{2} O \cdot H \cdot L_{\overline{V_1V_2}} \exp(-\frac{1}{2} x^{\gamma}) dx \\
        &= \frac{2^{\frac{1}{\gamma}} O \cdot S}{3} \int_{0}^{\infty} \exp(- x^{\gamma}) dx \\
        &= \frac{2^{\frac{1}{\gamma}} O \cdot S}{3} \frac{1}{\gamma} \Gamma(\frac{1}{\gamma}) \\
    \end{aligned}
\end{equation}
where $H$ is distance from $C$ to $\overline{V_1V_2}$, $L_{\overline{V_1V_2}}$ is the length of $\overline{V_1V_2}$, and $S=\frac{3}{2} L_{\overline{V_1V_2}} H$ is the area of $\triangle_{V_1V_2V_3}$.
The Euler's gamma function $\Gamma$ is defined as:
$$\Gamma(x) = \int_{0}^{\infty} t^{x-1} e^{-t} dt .$$
Since the above integration is the same for all three subregions of the triangle plane, we have:
\begin{equation}
    I(\gamma) = 3 I_{\overline{CV_1}\vee\overline{CV_2}}(\gamma) = O \cdot S \cdot \frac{2^{\frac{1}{\gamma}} \Gamma(\frac{1}{\gamma})}{\gamma} .
\end{equation}

\section{Additional Experimental Results}

\subsection{Novel View Synthesis}

We provide the detailed results of our experiments on each scene of the MipNeRF360~\cite{barron2022mipnerf360}, NeRF-Synthetic~\cite{mildenhall2020nerf}, Tanks and Temples~\cite{Knapitsch2017TanksAndTemples}, and Deep Blending~\cite{Hedman2018DeepBlending} datasets.
The experiments are described in Section \ref{sec:exp_cat1}.

\begin{figure*}[ht]
    \centering
    \includegraphics[width=0.95\linewidth]{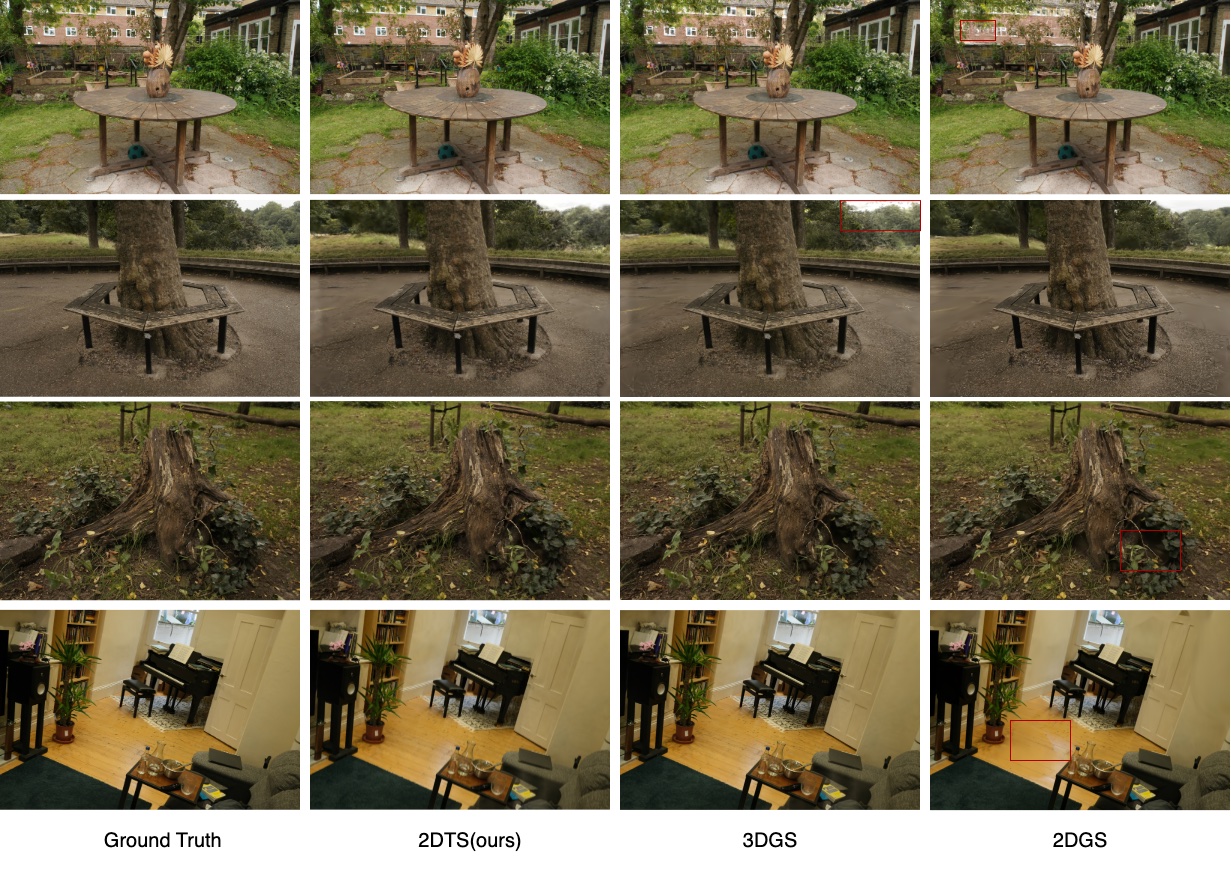}
    \caption{
        Visual comparison of the radiance field reconstruction between our method, 3DGS, and 2DGS on the MipNeRF360 dataset.
    }
    \label{fig:nvs-mipnerf360}
\end{figure*}

Figure \ref{fig:nvs-mipnerf360} shows qualitative comparisons on the MipNeRF360 dataset.
Tables \ref{tab:MipNeRF360-detail}, \ref{tab:NeRF-Synthetic-detail}, and \ref{tab:TT&DB-detail} show the quantitative results for the MipNeRF360, NeRF-Synthetic, Tanks and Temples, and Deep Blending datasets, respectively.
\textcolor{1st}{Red} indicates the best score in these tables while \textcolor{2nd}{orange} indicates the second best score.

\begin{table}[ht]
    \centering
    \small
    \resizebox{0.99\columnwidth}{!}{
    \begin{tabular}{c|l|c c c c c|c c c c|c}
        &            & \multicolumn{5}{c|}{Outdoor Scene}                                                                                         & \multicolumn{4}{c|}{Indoor scene}                                                                                          \\
        &            & bicycle                & flowers                & garden                 & stump                  & treehill               & room                   & counter                & kitchen                & bonsai                 & mean                   \\
        \hline
        \multirow{4}{*}{PSNR} 
        & M-Nerf360  & {24.37}                & \textcolor{2nd}{21.73} & \textcolor{2nd}{26.98} & {26.40}                & \textcolor{2nd}{22.87} & \textcolor{2nd}{31.63} & \textcolor{2nd}{29.55} & \textcolor{1st}{32.23} & \textcolor{1st}{33.46} & \textcolor{2nd}{27.69} \\
        & 3DGS       & {24.71}                & {21.09}                & {26.63}                & \textcolor{2nd}{26.45} & {22.33}                & {30.50}                & {29.07}                & \textcolor{2nd}{31.13} & {32.26}                & {27.24}                \\
        & 2DGS       & \textcolor{2nd}{24.82} & {20.99}                & {26.91}                & {26.41}                & {22.52}                & {30.86}                & {28.45}                & {30.62}                & {31.64}                & {27.03}                \\
        & 2DTS(Ours) & \textcolor{1st}{26.29} & \textcolor{1st}{22.54} & \textcolor{1st}{28.59} & \textcolor{1st}{26.89} & \textcolor{1st}{22.92} & \textcolor{1st}{32.56} & \textcolor{1st}{29.69} & {30.83}                & \textcolor{2nd}{33.27} & \textcolor{1st}{28.18} \\
        \hline
        \multirow{4}{*}{SSIM}
        & M-Nerf360  & {0.685}                & \textcolor{2nd}{0.583} & {0.813}                & {0.744}                & \textcolor{2nd}{0.632} & {0.913}                & {0.894}                & {0.920}                & \textcolor{2nd}{0.941} & {0.792}                \\
        & 3DGS       & {0.729}                & {0.571}                & {0.834}                & {0.762}                & {0.627}                & \textcolor{2nd}{0.922} & \textcolor{2nd}{0.913} & \textcolor{2nd}{0.922} & {0.926}                & {0.797}                \\
        & 2DGS       & \textcolor{2nd}{0.731} & {0.573}                & \textcolor{2nd}{0.845} & \textcolor{2nd}{0.764} & {0.630}                & {0.918}                & {0.908}                & \textcolor{1st}{0.927} & {0.940}                & \textcolor{2nd}{0.804} \\
        & 2DTS(Ours) & \textcolor{1st}{0.798} & \textcolor{1st}{0.667} & \textcolor{1st}{0.890} & \textcolor{1st}{0.787} & \textcolor{1st}{0.680} & \textcolor{1st}{0.946} & \textcolor{1st}{0.929} & {0.920}                & \textcolor{1st}{0.959} & \textcolor{1st}{0.842} \\
        \hline
        \multirow{4}{*}{LPIPS} 
        & M-Nerf360  & {0.301}                & \textcolor{2nd}{0.344} & {0.170}                & \textcolor{2nd}{0.261} & \textcolor{2nd}{0.339} & \textcolor{1st}{0.211} & \textcolor{2nd}{0.204} & \textcolor{2nd}{0.127} & \textcolor{1st}{0.176} & \textcolor{2nd}{0.237} \\
        & 3DGS       & \textcolor{2nd}{0.265} & {0.377}                & {0.147}                & {0.266}                & {0.362}                & {0.231}                & {0.212}                & {0.138}                & {0.214}                & {0.246}                \\
        & 2DGS       & {0.271}                & {0.378}                & \textcolor{2nd}{0.138} & {0.263}                & {0.369}                & \textcolor{2nd}{0.214} & \textcolor{1st}{0.197} & \textcolor{1st}{0.125} & {0.194}                & {0.239}                \\
        & 2DTS(Ours) & \textcolor{1st}{0.211} & \textcolor{1st}{0.290} & \textcolor{1st}{0.107} & \textcolor{1st}{0.244} & \textcolor{1st}{0.335} & {0.229}                & {0.206}                & {0.150}                & \textcolor{2nd}{0.192} & \textcolor{1st}{0.218} \\
    \end{tabular}
    }
    \caption{
        PSNR $\uparrow$, SSIM $\uparrow$, LPIPS $\downarrow$ scores for the MipNeRF360~\cite{barron2022mipnerf360} dataset.
    }
    \label{tab:MipNeRF360-detail}
\end{table}

\begin{table}[ht]
    \centering
    \resizebox{0.99\columnwidth}{!}{
    \begin{tabular}{c|l|c c c c c c c c|c}
        &            & chair                  & drums                  & ficus                  & hotdog                 & lego                   & materials              & mic                    & ship                   & mean                   \\
        \hline
        \multirow{4}{*}{PSNR}
        & Mip-NeRF   & {35.08}                & {25.26}                & {33.44}                & {37.38}                & {35.46}                & \textcolor{1st}{30.63} & \textcolor{1st}{36.38} & {30.46}                & {33.05}                \\
        & 3DGS       & \textcolor{1st}{35.87} & \textcolor{2nd}{26.15} & {34.90}                & \textcolor{1st}{37.70} & \textcolor{2nd}{35.90} & {30.12}                & \textcolor{2nd}{35.89} & \textcolor{1st}{31.07} & \textcolor{2nd}{33.45} \\
        & 2DGS       & {35.23}                & {26.11}                 & \textcolor{2nd}{35.37} & {37.35}                & {35.12}                & {29.69}                & {35.16}                & {30.56}                & {33.07}                \\
        & 2DTS(Ours) & \textcolor{2nd}{35.57} & \textcolor{1st}{26.23} & \textcolor{1st}{35.86} & \textcolor{2nd}{37.69} & \textcolor{1st}{36.03} & \textcolor{2nd}{30.19} & {35.80}                & \textcolor{2nd}{30.71} & \textcolor{1st}{33.51} \\
        \hline
        \multirow{4}{*}{SSIM}
        & Mip-NeRF   & {0.980}                & {0.934}                & {0.981}                & {0.982}                & {0.978}                & {0.959}                & {0.991}                & {0.885}                & {0.961}                \\
        & 3DGS       & \textcolor{1st}{0.988} & \textcolor{1st}{0.955} & {0.987}                & \textcolor{2nd}{0.985} & \textcolor{1st}{0.983} & \textcolor{1st}{0.962} & \textcolor{1st}{0.992} & \textcolor{1st}{0.907} & \textcolor{1st}{0.970} \\
        & 2DGS       & {0.985}                & \textcolor{2nd}{0.954} & \textcolor{2nd}{0.988} & {0.984}                & {0.980}                & {0.958}                & {0.991}                & {0.904}                & {0.968}                \\
        & 2DTS(Ours) & \textcolor{1st}{0.988} & \textcolor{2nd}{0.954} & \textcolor{1st}{0.990} & \textcolor{1st}{0.986} & \textcolor{1st}{0.983} & \textcolor{2nd}{0.961} & \textcolor{1st}{0.992} & \textcolor{2nd}{0.905} & \textcolor{1st}{0.970} \\
        \hline
        \multirow{4}{*}{LPIPS}
        & Mip-NeRF   & {0.041}                & {0.104}                & {0.045}                & {0.038}                & {0.053}                & {0.054}                & {0.024}                & {0.177}                & {0.067}                \\
        & 3DGS       & \textcolor{1st}{0.012} & \textcolor{1st}{0.036} & \textcolor{1st}{0.012} & \textcolor{1st}{0.020} & \textcolor{1st}{0.015} & \textcolor{1st}{0.033} & \textcolor{1st}{0.006} & \textcolor{1st}{0.105} & \textcolor{1st}{0.030} \\
        & 2DGS       & {0.015}                & \textcolor{2nd}{0.040} & \textcolor{1st}{0.012} & \textcolor{2nd}{0.023} & {0.020}                & \textcolor{2nd}{0.039} & \textcolor{2nd}{0.007} & \textcolor{2nd}{0.110} & \textcolor{2nd}{0.033} \\
        & 2DTS(Ours) & \textcolor{2nd}{0.014} & {0.047}                & \textcolor{1st}{0.012} & {0.024}                & \textcolor{2nd}{0.018} & {0.047}                & {0.008}                & {0.122}                & {0.037}                \\
    \end{tabular}
    }
    \caption{
        PSNR $\uparrow$, SSIM $\uparrow$, LPIPS $\downarrow$ scores for the NeRF-Synthetic~\cite{mildenhall2020nerf} dataset.
    }
    \label{tab:NeRF-Synthetic-detail}
\end{table}

\begin{table}[ht]
    \centering
    \resizebox{0.99\columnwidth}{!}{
    \begin{tabular}{c|l|c c|c c}
        &            & Truck                  & Train                  & Dr Johnson             & Playroom               \\
        \hline
        \multirow{4}{*}{PSNR}
        & M-Nerf360  & {24.91}                & {19.52}                & \textcolor{1st}{29.14} & {29.66}                \\
        & 3DGS       & \textcolor{2nd}{25.19} & {21.10}                & {28.77}                & \textcolor{2nd}{30.04} \\
        & 2DGS       & 25.10                  & \textcolor{2nd}{21.13} & {28.95}                & \textcolor{1st}{30.15} \\
        & 2DTS(Ours) & \textcolor{1st}{25.40} & \textcolor{1st}{21.38} & \textcolor{2nd}{28.98} & {29.76}                \\
        \hline
        \multirow{4}{*}{SSIM}
        & M-Nerf360  & {0.857}                & {0.660}                & \textcolor{2nd}{0.901} & {0.900}                \\
        & 3DGS       & \textcolor{2nd}{0.879} & \textcolor{2nd}{0.802} & {0.899}                & \textcolor{2nd}{0.906} \\
        & 2DGS       & 0.873                  & 0.790                  & {0.900}                & \textcolor{2nd}{0.906} \\
        & 2DTS(Ours) & \textcolor{1st}{0.886} & \textcolor{1st}{0.820} & \textcolor{1st}{0.906} & \textcolor{1st}{0.909} \\
        \hline
        \multirow{4}{*}{LPIPS}
        & M-Nerf360  & \textcolor{2nd}{0.159} & {0.354}                & \textcolor{1st}{0.237} & \textcolor{2nd}{0.252} \\
        & 3DGS       & \textcolor{1st}{0.148} & \textcolor{1st}{0.218} & \textcolor{2nd}{0.244} & \textcolor{1st}{0.241} \\
        & 2DGS       & {0.173}                & {0.251}                & 0.256                  & {0.257}                \\
        & 2DTS(Ours) & {0.160}                & \textcolor{2nd}{0.247} & {0.310}                & {0.311}                \\
        \end{tabular}
    }
    \caption{
        PSNR $\uparrow$, SSIM $\uparrow$, LPIPS $\downarrow$ scores for the Tanks \& Temples~\cite{Knapitsch2017TanksAndTemples} and Deep Blending~\cite{Hedman2018DeepBlending} datasets. 
    }
    \label{tab:TT&DB-detail}
\end{table}

\paragraph{Additional NVS Clarifications.}
We additionally report efficiency on Mip-NeRF360 and ablations on the NVS setting.
All results in Table~\ref{tab:opacity_ablation_suppl} use the same protocol as the main paper unless otherwise specified.

\begin{table}[ht]
    \centering
    \small
    \resizebox{0.95\columnwidth}{!}{
    \begin{tabular}{l|c c c}
                    & Training Time (min)$\downarrow$ & Memory (GB)$\downarrow$ & Rendering FPS$\uparrow$ \\
        \hline
        M-NeRF360   & 2735                            & \textcolor{1st}{0.38}   & 0.08                    \\
        3DGS        & \textcolor{2nd}{60}             & 9.32                    & \textcolor{1st}{87}     \\
        2DGS        & 98                              & \textcolor{2nd}{2.60}   & 32                      \\
        2DTS (Ours) & \textcolor{1st}{48}             & 2.63                    & \textcolor{2nd}{50}     \\
    \end{tabular}
    }
    \caption{Training time, memory, and rendering FPS on Mip-NeRF360.}
    \label{tab:rendering_speed_suppl}
\end{table}

\begin{table}[ht]
    \centering
    \small
    \resizebox{0.9\columnwidth}{!}{
    \begin{tabular}{l|c c c}
                        & PSNR$\uparrow$         & SSIM$\uparrow$         & LPIPS$\downarrow$      \\
        \hline
        Full Model      & 28.18                  & 0.842                  & 0.218                  \\
        Vertex Color    & 28.39                  & 0.844                  & 0.209                  \\
        Uniform Opacity & 26.51                  & 0.781                  & 0.303                  \\
        Pre-downsampled & 27.32                  & 0.814                  & 0.246                  \\
    \end{tabular}
    }
    \caption{Ablations on the NVS setting of Mip-NeRF360. The main-paper setting uses face color and runtime bilinear downsampling.}
    \label{tab:opacity_ablation_suppl}
\end{table}

\subsection{Mesh Reconstruction}

\begin{figure*}[ht]
    \centering
    \includegraphics[width=0.95\linewidth]{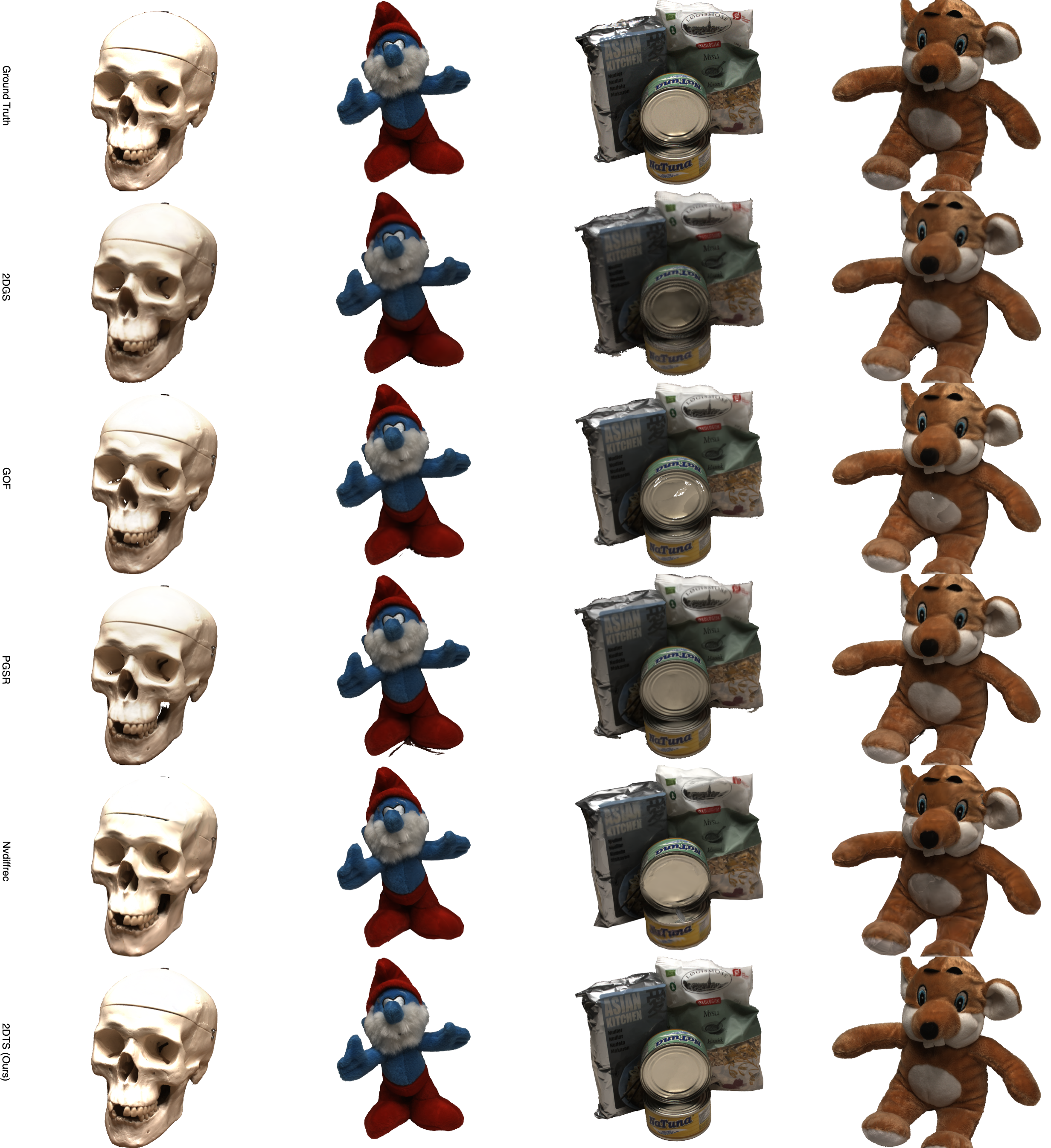}
    \caption{
        Visual comparison of the mesh reconstruction between our method, 2DGS, GOF, PGSR, and Nvdiffrec on the DTU dataset.
    }
    \label{fig:mesh-compare-dtu}
\end{figure*}

We provide the detailed results of our experiments on each scene of the NeRF-Synthetic~\cite{mildenhall2020nerf} and DTU~\cite{jensen2014large} datasets.
The experiments are described in Section \ref{sec:exp_cat2}.
Figure \ref{fig:mesh-compare-dtu} shows qualitative comparisons on the DTU dataset.
Table \ref{tab:NeRF-synthetic-mesh-detail} shows the results for the NeRF-Synthetic dataset.
Table \ref{tab:DTU-mesh-detail} shows the results for the DTU dataset.

For a more apples-to-apples comparison of rendering-based metrics, Table~\ref{tab:kaolin_rendering_suppl} reports results when all methods are rendered with the Kaolin renderer.

\begin{table*}[ht]
    \centering
    \small
    \resizebox{0.95\linewidth}{!}{
    \begin{tabular}{l|c c c c|c c c c}
                         & \multicolumn{4}{c|}{\makecell{NeRF-Synthetic}}                                                          & \multicolumn{4}{c}{\makecell{DTU}}                                                      \\
                         & PSNR$\uparrow$         & SSIM$\uparrow$         & LPIPS$\downarrow$      & Count(K)$\downarrow$ & PSNR$\uparrow$         & SSIM$\uparrow$         & LPIPS$\downarrow$      & Count(K)$\downarrow$ \\
        \hline
        2DGS-Mesh        & 21.83                  & 0.866                  & 0.145                  & 489                  & 21.50                  & 0.851                  & 0.227                  & \textcolor{2nd}{268} \\
        GOF-Mesh         & \textcolor{2nd}{22.13} & 0.876                  & 0.136                  & 501                  & 22.25                  & 0.869                  & \textcolor{2nd}{0.193} & 1062                 \\
        PGSR-Mesh        & 21.71                  & \textcolor{2nd}{0.877} & \textcolor{2nd}{0.134} & 361                  & \textcolor{2nd}{22.73} & \textcolor{1st}{0.875} & \textcolor{1st}{0.190} & 1047                 \\
        Nvdiffrec        & 18.93                  & 0.844                  & 0.151                  & \textcolor{2nd}{83}  & 16.49                  & 0.760                  & 0.296                  & 567                  \\
        2DTS-Mesh (Ours) & \textcolor{1st}{25.60} & \textcolor{1st}{0.909} & \textcolor{1st}{0.118} & \textcolor{1st}{73}  & \textcolor{1st}{24.22} & \textcolor{2nd}{0.874} & 0.212                  & \textcolor{1st}{265} \\
    \end{tabular}
    }
    \caption{Rendering-based metrics with Kaolin for all methods.}
    \label{tab:kaolin_rendering_suppl}
\end{table*}

\begin{table*}[ht]
    \centering
    \resizebox{0.95\linewidth}{!}{
    \begin{tabular}{c|l|c c c c c c c c|c}
        &                  & chair                  & drums                  & ficus                  & hotdog                 & lego                   & materials              & mic                    & ship                   & mean                   \\
        \hline
        \multirow{5}{*}{PSNR}
        & 2DGS-Mesh        & 25.64                  & 19.73                  & 18.01                  & 25.91                  & 22.72                  & 18.83                  & 22.13                  & 21.64                  & 21.83                  \\
        & GOF-Mesh         & 25.73                  & 19.01                  & 18.86                  & 25.43                  & 23.98                  & 20.68                  & 21.72                  & 21.61                  & 22.13                  \\
        & PGSR-Mesh        & 25.99                  & 18.38                  & 20.57                  & 24.25                  & 23.58                  & 18.00                  & 20.64                  & 22.26                  & 21.71                  \\
        & Nvdiffrec        & \textcolor{2nd}{31.68} & \textcolor{2nd}{24.32} & \textcolor{2nd}{29.92} & \textcolor{2nd}{32.87} & \textcolor{2nd}{28.94} & \textcolor{2nd}{26.52} & \textcolor{2nd}{30.12} & \textcolor{2nd}{25.71} & \textcolor{2nd}{28.76} \\
        & 2DTS-Mesh (Ours) & \textcolor{1st}{34.16} & \textcolor{1st}{25.12} & \textcolor{1st}{30.37} & \textcolor{1st}{35.01} & \textcolor{1st}{33.17} & \textcolor{1st}{27.03} & \textcolor{1st}{32.69} & \textcolor{1st}{28.87} & \textcolor{1st}{30.80} \\
        \hline
        \multirow{5}{*}{SSIM} 
        & 2DGS-Mesh        & 0.898                  & 0.844                  & 0.876                  & 0.920                  & 0.852                  & 0.847                  & 0.915                  & 0.772                  & 0.866                  \\
        & GOF-Mesh         & 0.906                  & 0.845                  & 0.886                  & 0.933                  & 0.879                  & 0.865                  & 0.916                  & 0.782                  & 0.876                  \\
        & PGSR-Mesh        & 0.909                  & 0.847                  & 0.896                  & 0.932                  & 0.870                  & 0.850                  & 0.915                  & 0.795                  & 0.877                  \\
        & Nvdiffrec        & \textcolor{2nd}{0.968} & \textcolor{2nd}{0.920} & \textcolor{2nd}{0.966} & \textcolor{2nd}{0.972} & \textcolor{2nd}{0.945} & \textcolor{1st}{0.932} & \textcolor{2nd}{0.974} & \textcolor{2nd}{0.826} & \textcolor{2nd}{0.938} \\
        & 2DTS-Mesh (Ours) & \textcolor{1st}{0.982} & \textcolor{1st}{0.944} & \textcolor{1st}{0.973} & \textcolor{1st}{0.976} & \textcolor{1st}{0.972} & \textcolor{2nd}{0.928} & \textcolor{1st}{0.986} & \textcolor{1st}{0.879} & \textcolor{1st}{0.955} \\
        \hline
        \multirow{5}{*}{LPIPS}
        & 2DGS-Mesh        & 0.101                  & 0.161                  & 0.127                  & 0.109                  & 0.151                  & 0.149                  & 0.088                  & 0.271                  & 0.145                  \\
        & GOF-Mesh         & 0.096                  & 0.161                  & 0.118                  & 0.103                  & 0.128                  & 0.134                  & 0.091                  & 0.258                  & 0.136                  \\
        & PGSR-Mesh        & 0.090                  & 0.159                  & 0.102                  & 0.098                  & 0.136                  & 0.148                  & 0.090                  & 0.253                  & 0.134                  \\
        & Nvdiffrec        & \textcolor{2nd}{0.045} & \textcolor{2nd}{0.092} & \textcolor{2nd}{0.050} & \textcolor{1st}{0.063} & \textcolor{2nd}{0.065} & \textcolor{1st}{0.089} & \textcolor{2nd}{0.042} & \textcolor{2nd}{0.194} & \textcolor{2nd}{0.080} \\
        & 2DTS-Mesh (Ours) & \textcolor{1st}{0.025} & \textcolor{1st}{0.067} & \textcolor{1st}{0.044} & \textcolor{1st}{0.063} & \textcolor{1st}{0.038} & \textcolor{2nd}{0.100} & \textcolor{1st}{0.025} & \textcolor{1st}{0.165} & \textcolor{1st}{0.066} \\
        \hline
        \multirow{5}{*}{CD(1e-3)}
        & 2DGS-Mesh        & {72.4}                 & {44.8}                 & {80.6}                 & {81.6}                 & \textcolor{2nd}{15.4}   & \textcolor{1st}{9.6}  & {27.3}                 & \textcolor{2nd}{48.8}  & \textcolor{2nd}{47.5}  \\
        & GOF-Mesh         & {79.5}                 & \textcolor{2nd}{26.9}  & {15.7}                 & {118.6}                & {46.3}                  & {11.3}                & {12.4}                 & {429.4}                & {92.9}                 \\
        & PGSR-Mesh        & {113.1}                & {28.1}                 & {23.9}                 & {122.0}                & {34.8}                  & {17.2}                & {19.7}                 & \textcolor{1st}{35.1}  & {49.2}                 \\
        & Nvdiffrec        & \textcolor{2nd}{57.4}  & {32.5}                 & \textcolor{2nd}{15.4}  & \textcolor{2nd}{27.2}  & {26.7}                  & {18.0}                & \textcolor{1st}{9.8}   & {393.0}                & {72.5}                 \\
        & 2DTS-Mesh (Ours) & \textcolor{1st}{23.0}  & \textcolor{1st}{25.8}  & \textcolor{1st}{8.5}   & \textcolor{1st}{14.6}  & \textcolor{1st}{4.7}    & \textcolor{2nd}{9.8}  & \textcolor{2nd}{11.8}  & {124.2}                & \textcolor{1st}{27.8}  \\
        \hline
        \multirow{5}{*}{Count(K)}
        & 2DGS-Mesh        & 355                    & 480                    & 119                    & 414                    & 758                    & 485                    & 235                    & 1065                   & 489                    \\
        & GOF-Mesh         & 464                    & 465                    & 152                    & 366                    & 635                    & 509                    & 210                    & 1205                   & 501                    \\
        & PGSR-Mesh        & 273                    & 430                    & 221                    & 300                    & 534                    & 431                    & 156                    & 545                    & 361                    \\
        & Nvdiffrec        & \textcolor{2nd}{104}   & \textcolor{1st}{68}    & \textcolor{2nd}{41}    & \textcolor{2nd}{57}    & \textcolor{2nd}{116}   & \textcolor{1st}{65}    & \textcolor{1st}{24}    & \textcolor{2nd}{187}   & \textcolor{2nd}{83}    \\
        & 2DTS-Mesh (Ours) & \textcolor{1st}{84}    & \textcolor{2nd}{77}    & \textcolor{1st}{39}    & \textcolor{1st}{53}    & \textcolor{1st}{104}   & \textcolor{2nd}{73}    & \textcolor{2nd}{72}    & \textcolor{1st}{85}    & \textcolor{1st}{73}    \\
    \end{tabular}
    }
    \caption{
        PSNR $\uparrow$, SSIM $\uparrow$, LPIPS $\downarrow$ and Chamfer Distance (CD) $\downarrow$ scores for the NeRF-Synthetic~\cite{mildenhall2020nerf} dataset.
        Face counts of the meshes are also reported.
    }
    \label{tab:NeRF-synthetic-mesh-detail}
\end{table*}

\begin{table*}[ht]
    \centering
    \resizebox{0.95\linewidth}{!}{
    \begin{tabular}{c|l|c c c c c c c c c c c c c c c|c}
        &                  & 24                     & 37                     & 40                     & 55                     & 63                     & 65                     & 69                     & 83                     & 97                     & 105                    & 106                    & 110                    & 114                    & 118                    & 122                    & mean                   \\
        \hline
        \multirow{5}{*}{PSNR}
        & 2DGS-Mesh        & 21.36                  & 20.61                  & 22.69                  & 23.05                  & 15.80                  & 22.36                  & 19.85                  & 25.24                  & 20.39                  & 23.79                  & 21.09                  & 16.96                  & 21.14                  & 22.56                  & 25.65                  & 21.50                  \\
        & GOF-Mesh         & 21.48                  & 20.95                  & 23.05                  & 23.16                  & 23.24                  & 22.62                  & 19.56                  & 24.90                  & 21.94                  & 24.20                  & 19.46                  & 20.02                  & 20.91                  & 22.80                  & 25.51                  & 22.25                  \\
        & PGSR-Mesh        & 22.12                  & 21.19                  & 23.43                  & 23.17                  & 25.31                  & 22.88                  & 19.92                  & 24.10                  & 21.57                  & 23.63                  & 21.98                  & 19.11                  & 23.73                  & 22.25                  & 26.52                  & 22.73                  \\
        & Nvdiffrec        & \textcolor{2nd}{23.34} & \textcolor{2nd}{23.75} & \textcolor{2nd}{24.70} & \textcolor{2nd}{25.42} & \textcolor{2nd}{29.83} & \textcolor{2nd}{26.00} & \textcolor{2nd}{23.59} & \textcolor{2nd}{28.77} & \textcolor{2nd}{24.66} & \textcolor{2nd}{27.34} & \textcolor{2nd}{23.86} & \textcolor{2nd}{24.55} & \textcolor{2nd}{25.67} & \textcolor{2nd}{26.13} & \textcolor{2nd}{30.77} & \textcolor{2nd}{25.89} \\
        & 2DTS-Mesh (Ours) & \textcolor{1st}{30.68} & \textcolor{1st}{28.31} & \textcolor{1st}{30.85} & \textcolor{1st}{30.57} & \textcolor{1st}{32.31} & \textcolor{1st}{29.13} & \textcolor{1st}{25.03} & \textcolor{1st}{29.62} & \textcolor{1st}{27.69} & \textcolor{1st}{29.81} & \textcolor{1st}{26.49} & \textcolor{1st}{26.44} & \textcolor{1st}{29.45} & \textcolor{1st}{30.13} & \textcolor{1st}{33.27} & \textcolor{1st}{29.32} \\
        \hline
        \multirow{5}{*}{SSIM} 
        & 2DGS-Mesh        & 0.716                  & 0.766                  & 0.691                  & 0.810                  & 0.898                  & 0.923                  & 0.864                  & 0.948                  & 0.852                  & 0.886                  & 0.864                  & 0.866                  & 0.848                  & 0.907                  & 0.929                  & 0.851                  \\
        & GOF-Mesh         & 0.763                  & 0.777                  & 0.735                  & 0.844                  & 0.921                  & 0.926                  & 0.873                  & 0.949                  & 0.870                  & 0.893                  & 0.878                  & 0.884                  & 0.863                  & 0.917                  & 0.936                  & 0.869                  \\
        & PGSR-Mesh        & \textcolor{2nd}{0.777} & \textcolor{2nd}{0.793} & \textcolor{2nd}{0.739} & \textcolor{2nd}{0.846} & \textcolor{2nd}{0.932} & \textcolor{2nd}{0.931} & 0.878                  & 0.949                  & 0.876                  & 0.895                  & \textcolor{2nd}{0.888} & 0.881                  & \textcolor{2nd}{0.877} & \textcolor{2nd}{0.922} & 0.940                  & \textcolor{2nd}{0.875} \\
        & Nvdiffrec        & 0.703                  & 0.762                  & 0.658                  & 0.833                  & 0.928                  & 0.925                  & \textcolor{2nd}{0.888} & \textcolor{2nd}{0.957} & \textcolor{2nd}{0.879} & \textcolor{2nd}{0.896} & 0.865                  & \textcolor{2nd}{0.902} & 0.837                  & {0.921}                & \textcolor{2nd}{0.948} & 0.860                  \\
        & 2DTS-Mesh (Ours) & \textcolor{1st}{0.925} & \textcolor{1st}{0.918} & \textcolor{1st}{0.913} & \textcolor{1st}{0.958} & \textcolor{1st}{0.965} & \textcolor{1st}{0.961} & \textcolor{1st}{0.915} & \textcolor{1st}{0.970} & \textcolor{1st}{0.935} & \textcolor{1st}{0.946} & \textcolor{1st}{0.947} & \textcolor{1st}{0.937} & \textcolor{1st}{0.936} & \textcolor{1st}{0.962} & \textcolor{1st}{0.971} & \textcolor{1st}{0.944} \\
        \hline
        \multirow{5}{*}{LPIPS}
        & 2DGS-Mesh        & 0.374                  & 0.231                  & 0.371                  & 0.217                  & 0.164                  & 0.161                  & 0.278                  & 0.121                  & 0.231                  & 0.218                  & 0.234                  & 0.221                  & 0.262                  & 0.182                  & 0.139                  & 0.227                  \\
        & GOF-Mesh         & 0.280                  & 0.206                  & \textcolor{2nd}{0.307} & \textcolor{2nd}{0.169} & 0.132                  & 0.143                  & {0.252}                & 0.113                  & 0.192                  & {0.195}                & 0.201                  & 0.200                  & 0.229                  & 0.160                  & 0.119                  & 0.193                  \\
        & PGSR-Mesh        & \textcolor{2nd}{0.277} & \textcolor{2nd}{0.203} & 0.316                  & 0.174                  & \textcolor{2nd}{0.118} & \textcolor{2nd}{0.141} & \textcolor{1st}{0.245} & 0.114                  & \textcolor{2nd}{0.188} & \textcolor{2nd}{0.191} & \textcolor{2nd}{0.194} & \textcolor{2nd}{0.198} & \textcolor{2nd}{0.218} & \textcolor{2nd}{0.159} & \textcolor{2nd}{0.117} & \textcolor{2nd}{0.190} \\
        & Nvdiffrec        & 0.404                  & 0.260                  & 0.415                  & 0.226                  & 0.150                  & 0.159                  & 0.260                  & \textcolor{2nd}{0.107} & 0.213                  & 0.220                  & 0.247                  & 0.208                  & 0.263                  & 0.176                  & 0.118                  & 0.228                  \\
        & 2DTS-Mesh (Ours) & \textcolor{1st}{0.141} & \textcolor{1st}{0.135} & \textcolor{1st}{0.209} & \textcolor{1st}{0.131} & \textcolor{1st}{0.089} & \textcolor{1st}{0.104} & \textcolor{2nd}{0.249} & \textcolor{1st}{0.095} & \textcolor{1st}{0.162} & \textcolor{1st}{0.157} & \textcolor{1st}{0.148} & \textcolor{1st}{0.177} & \textcolor{1st}{0.153} & \textcolor{1st}{0.127} & \textcolor{1st}{0.094} & \textcolor{1st}{0.145} \\
        \hline
        \multirow{5}{*}{CD}
        & 2DGS-Mesh        & {0.603}                & {0.825}                & \textcolor{1st}{0.332} & \textcolor{2nd}{0.351} & {0.956}                & {0.871}                & {0.798}                & {1.330}                & \textcolor{2nd}{1.188} & {0.675}                & {0.654}                & {2.206}                & {0.393}                & {0.676}                & {0.473}                & {0.819}                \\
        & GOF-Mesh         & {0.496}                & {0.836}                & {0.461}                & {0.387}                & {1.409}                & {0.812}                & {0.809}                & {1.274}                & {1.317}                & {0.665}                & {0.733}                & {1.291}                & {0.486}                & {0.692}                & {0.506}                & {0.812}                \\
        & PGSR-Mesh        & \textcolor{1st}{0.367} & \textcolor{1st}{0.544} & {0.400}                & \textcolor{1st}{0.336} & \textcolor{2nd}{0.779} & \textcolor{1st}{0.576} & \textcolor{1st}{0.488} & \textcolor{2nd}{1.080} & \textcolor{1st}{0.635} & \textcolor{2nd}{0.585} & \textcolor{1st}{0.463} & \textcolor{1st}{0.548} & \textcolor{1st}{0.304} & \textcolor{1st}{0.369} & \textcolor{1st}{0.340} & \textcolor{1st}{0.521} \\
        & Nvdiffrec        & {2.436}                & {1.701}                & {2.438}                & {1.351}                & {2.573}                & {1.733}                & {1.568}                & {1.822}                & {2.042}                & {1.480}                & {1.649}                & {3.464}                & {1.732}                & {1.451}                & {1.248}                & {1.907}                \\
        & 2DTS-Mesh (Ours) & \textcolor{2nd}{0.458} & \textcolor{2nd}{0.593} & \textcolor{2nd}{0.360} & {0.395}                & \textcolor{1st}{0.632} & \textcolor{2nd}{0.752} & \textcolor{2nd}{0.591} & \textcolor{1st}{0.848} & {1.197}                & \textcolor{1st}{0.373} & \textcolor{2nd}{0.482} & \textcolor{2nd}{0.764} & \textcolor{2nd}{0.349} & \textcolor{2nd}{0.401} & \textcolor{2nd}{0.361} & \textcolor{2nd}{0.570} \\
        \hline
        \multirow{5}{*}{Count(K)}
        & 2DGS-Mesh        & \textcolor{1st}{492}   & \textcolor{1st}{453}   & \textcolor{1st}{332}   & \textcolor{1st}{257}   & \textcolor{2nd}{260}   & \textcolor{2nd}{349}   & \textcolor{2nd}{304}   & 207                    & \textcolor{2nd}{182}   & \textcolor{2nd}{216}   & \textcolor{2nd}{229}   & \textcolor{2nd}{175}   & \textcolor{1st}{140}   & \textcolor{2nd}{212}   & \textcolor{2nd}{205}   & \textcolor{2nd}{268}   \\
        & GOF-Mesh         & 1526                   & 1823                   & 1278                   & 954                    & 1722                   & 1302                   & 1129                   & 792                    & 806                    & 852                    & 850                    & 756                    & 551                    & 788                    & 798                    & 1062                   \\
        & PGSR-Mesh        & 1475                   & 1509                   & 1320                   & 958                    & 1579                   & 1371                   & 1183                   & 847                    & 705                    & 869                    & 930                    & 678                    & 568                    & 875                    & 834                    & 1047                   \\
        & Nvdiffrec        & 2056                   & 1007                   & 1037                   & \textcolor{2nd}{418}   & 615                    & 534                    & 483                    & \textcolor{2nd}{155}   & 429                    & 243                    & 346                    & 257                    & 392                    & 322                    & 208                    & 567                    \\
        & 2DTS-Mesh (Ours) & \textcolor{2nd}{535}   & \textcolor{2nd}{508}   & \textcolor{2nd}{717}   & 531                    & \textcolor{1st}{183}   & \textcolor{1st}{180}   & \textcolor{1st}{53}    & \textcolor{1st}{67}    & \textcolor{1st}{118}   & \textcolor{1st}{215}   & \textcolor{1st}{209}   & \textcolor{1st}{71}    & \textcolor{2nd}{270}   & \textcolor{1st}{161}   & \textcolor{1st}{162}   & \textcolor{1st}{265}   \\
    \end{tabular}
    }
    \caption{
        PSNR $\uparrow$, SSIM $\uparrow$, LPIPS $\downarrow$ and Chamfer Distance (CD) $\downarrow$ scores for the DTU~\cite{jensen2014large} dataset.
        Face counts of the meshes are also reported.
    }
    \label{tab:DTU-mesh-detail}
\end{table*}

\end{document}